\documentclass[letterpaper, 10 pt, conference]{ieeeconf}  

\IEEEoverridecommandlockouts                              

\usepackage{graphics} 
\usepackage{booktabs}
\usepackage{epsfig} 
\usepackage{mathptmx} 
\usepackage{times} 
\usepackage{amsmath} 
\usepackage{amssymb}  
\usepackage{multirow}
\usepackage[ruled,vlined]{algorithm2e}
\usepackage{xcolor}
\usepackage{stackengine} 
\usepackage{graphicx}
\usepackage{subcaption}
\usepackage{caption}
\title{\LARGE \bf
LeARN: Learnable and Adaptive Representations for Nonlinear Dynamics in System Identification
}

\author{Arunabh Singh $^{1}$ and Joyjit Mukherjee$^{2}$
\thanks{$^{1}$ Author is with the Visual Computing Lab, Indian Institute of Science, Bengaluru, India.
{\tt\small arunabhsingh25@gmail.com}}%
\thanks{$^{2}$ Author is with the Department of Electrical and Electronics Engineering, BITS Pilani Hyderabad Campus, Hyderabad, India.
{\tt\small j.mukherjee@hyderabad.bits-pilani.ac.in}}%
}

\usepackage[T1]{fontenc}
\usepackage[utf8]{inputenc}

\begin{document}

\maketitle
\thispagestyle{empty}
\pagestyle{empty}

\begin{abstract}
System identification, the process of deriving mathematical models of dynamical systems from observed input-output data, has undergone a paradigm shift with the advent of learning-based methods. Addressing the intricate challenges of data-driven discovery in nonlinear dynamical systems,
these methods have garnered significant attention. Among them, Sparse Identification of Nonlinear Dynamics (SINDy) has emerged as a transformative approach, distilling complex dynamical
behaviors into interpretable linear combinations of basis functions. However, SINDy’s reliance on
domain-specific expertise to construct its foundational ’library’ of basis functions limits its adaptability and universality. In this work, we introduce a nonlinear system identification framework
LeARN that transcends the need for prior domain knowledge by learning the library of basis functions directly from data. To enhance adaptability to evolving system dynamics under varying noise
conditions, we employ a novel meta-learning-based system identification approach that utilizes a
light-weight Deep Neural Network (DNN) to dynamically refine these basis functions. This not
only captures intricate system behaviors but also adapts effectively to new dynamical regimes.
We validate our framework on the Neural Fly dataset, showcasing its robust adaptation and
generalization capabilities. Despite its simplicity, our LeARN achieves competitive dynamical error
performance to SINDy. This work presents a step towards autonomous discovery of dynamical
systems, paving the way for a future where machine learning uncovers the governing principles of
complex systems without requiring extensive domain-specific interventions.
\end{abstract}

\section{INTRODUCTION}

Robustness of robotic systems in unstructured environments requires adaptive control laws \cite{c1}, yet their efficacy is fundamentally contingent upon the accuracy of the underlying plant model. Historically, these models have relied on physics-based equations to ensure reliability and physical interpretability under ideal conditions. However, such models face critical limitations:
\begin{enumerate}
    \item Complex nonlinear dynamics: Many robotic systems operate in high-dimensional spaces with nonlinear interactions, making precise modeling difficult.
    \item Unknown or un-modeled dynamics: Environmental disturbances and uncertainties often introduce dynamics that are challenging to capture using predefined equations.
    \item High-dimensional inputs: Robots with numerous sensors and actuators generate large-scale data that physics-based models may struggle to incorporate.
\end{enumerate}

Traditional system identification typically relies on physics-based models~\cite{c4}, which, despite their theoretical grounding, are often constrained by a reliance on domain expertise and frequently fail to capture complex nonlinear behaviors and un-modeled environmental disturbances~\cite{c5}. Although early extensions like the Wiener and Volterra series~\cite{c9}  attempted to address this, the field has increasingly pivoted toward data-driven methods, with neural networks proving effective for identifying nonlinear systems~\cite{c6}. Modern deep learning approaches, such as universal approximators~\cite{c8} and Physics-Informed Neural Networks (PINNs)~\cite{c10}, excel in high-dimensional modeling, but lack the interpretability required for safety-critical robotics. In contrast, Sparse Identification of Nonlinear Dynamics (SINDy)~\cite{c2} ensures interpretability through sparsity, but remains limited by its dependence on predefined domain-specific basis libraries. This necessity for prior knowledge hinders scalability and flexibility when encountering unknown or evolving dynamics. Unlike previous autoencoder-based approaches~\cite{c16} that attempt to resolve this by learning latent coordinate transformations to fit fixed, human-designed libraries; although the requirement for domain expertise is effectively addressed there, the interpretability of the resulting learned dynamics equations is significantly diminished because of the latent space coordinate representation; we instead introduce \textbf{LeARN}. Our framework meta-learns the functional basis library itself directly from data in the physical state space, bridging the gap between deep learning's expressiveness and SINDy's interpretability while enabling rapid adaptation to unseen dynamics via Model-Agnostic Meta-Learning (MAML)~\cite{c12}. Our key contributions are:
\begin{enumerate}
    \item Unlike traditional SINDy, our method does not require a predefined function library, thus enhancing flexibility and eliminating the need for domain expertise.
    \item By meta-learning~\cite{c12}~\cite{c11} the library of basis functions, \textbf{LeARN} enables adaptability to changing dynamics and noisy conditions.
    \item  By employing light-weight deep neural networks, we retain the ability to model complex nonlinearities while the learned basis functions facilitate structural interpretability in terms of identifying the input features contributing dominantly to system dynamics.
\end{enumerate}

In this paper, we test our approach on the Neural Fly dataset~\cite{c3}, which provides a challenging benchmark for nonlinear system identification. Therefore, throughout the paper, our principal robotic system of interest for the demonstration of our framework will be a quadrotor.

\section{Preliminary}
\label{sec:Preliminary}

\subsection{Sparse System Identification}
SINDy~\cite{c2} (Sparse Identification for Nonlinear Dynamics) employs sparse regression~\cite{c17} to model system dynamics, identifying the terms that contribute prominently to the dynamical model. SINDy models the dynamics as follows:
\begin{equation}
    \dot{X} \approx \Theta(X)\mathcal{E}, 
\end{equation}
where $\Theta(\cdot)$ constitutes candidate basis functions. $\mathcal{E}$ is a matrix of coefficients for terms obtained by processing $\Theta(X)$, where $X$ represents our features of interest to model the dynamics. 

Choosing basis functions for $\Theta(\cdot)$ is a problem-specific and non-trivial task. Since the best choice is often not clear, we attempt to address the problem in this paper. Since our system of interest is a quadrotor, the features of interest $X$ comprise the states and control inputs as detailed in Sec~\ref{sec:Exp}. 



\subsection{Meta-Learning}
Meta-Learning~\cite{c11} is a mechanism of learning to learn. It seeks parameters that optimize a meta-objective over a distribution of tasks. Model Agnostic Meta-Learning (MAML)~\cite{c12} provides a gradient-based framework to leverage meta-learning approaches to different kinds of tasks while being agnostic to the learning framework and the architecture of the neural network model. For MAML, the bi-level meta-learning problem can be formulated as
\begin{equation}
\label{eq:meta_learning}
\begin{aligned}
\theta^* \in \arg\min_\theta 
& \left( \frac{1}{M} \sum_{i=1}^M \ell_i(\phi_i, \mathcal{D}_i^{\text{eval}}) + \mu_{\text{meta}} \|\theta\|_2^2 \right) \\
\text{s.t. } & \phi_i = \theta - \eta\nabla_{\theta}\ell_i(\theta,\mathcal{D}_i^{\text{train}})
\end{aligned},
\end{equation}
 where the optimal meta-parameters \( \theta^* \) minimize the average evaluation loss across \( M \) tasks. Each task \( i \) is associated with a task-specific loss function \( \ell_i \), a training dataset \( \mathcal{D}_i^{\text{train}} \), and an evaluation dataset \( \mathcal{D}_i^{\text{eval}} \). The task-specific parameters \( \phi_i \) are derived by applying a gradient-based adaptation mechanism which adapts the meta-parameters \( \theta \) to the task using the task's training dataset. Here, \( \mu_{\text{meta}} \geq 0 \) is the regularization coefficient. This formulation ensures that the meta-parameters \( \theta \) are well-suited for adaptation across a variety of tasks. Meta-learning based approaches have been applied to system identification 
in linear time-varying settings~\cite{lin2020sysid} and for Koopman-based 
modeling of nonlinear systems with parametric uncertainty~\cite{han2025mako}. 
They have also been used in adaptive control to model residual aerodynamical 
interactions~\cite{c3},~\cite{c18} for control.~\cite{c19} meta-learns an adaptive controller based on closed-loop tracking simulations. Our work differs in that we use meta-learning to learn the basis 
function library used in SINDy-style sparse identification, 
operating directly on the physical state and control variables 
rather than on a latent representation. As we will see in Sec.~\ref{sec:Method}, meta-learning provides an effective initialization for parameterized basis functions, ensuring rapid adaptation across varying system dynamics.

\section{Methodology}
\label{sec:Method}

\subsection{Problem Formulation}
Since our system of interest in this paper is a quadrotor, we consider the following dynamics. 
 \begin{align}
    \dot{\mathbf{p}} &= \mathbf{v}, \quad m\dot{\mathbf{v}} = m\mathbf{g} + R\mathbf{f}_u + \mathbf{f}_a, \label{eq:linear_dynamics} \\
    \dot{R} &= R{S}(\boldsymbol{\omega}), \quad J\dot{\boldsymbol{\omega}} = J\boldsymbol{\omega} \times \boldsymbol{\omega} + \boldsymbol{\tau}_u + \boldsymbol{\tau}_a, \label{eq:rotational_dynamics}
\end{align}
where the global position vector $p\in\mathbb{R}^3$, velocity vector $v\in\mathbb{R}^3$, attitude rotation matrix $R\in\mathbf{SO(3)}$ and body angular velocity $\omega\in\mathbb{R}^3$ give the states of the quadrotor. Furthermore, $m$ is the mass of the quadrotor, $J$ is the inertia matrix, $g$ is gravitational acceleration, $f_a$ is
the external disturbance force, $S(\cdot)$ is the skew-symmetric mapping, $\tau_u$ and $\tau_a$ are control and external disturbance torques, respectively~\cite{c13}.

We first model partial dynamics of the quadrotor for which we will use the following formulation of:
\begin{equation}
\label{eq:partial_dynamics}
    \dot{v} = f(v, u=Rf_u),
\end{equation}
where based on Eq~\eqref{eq:linear_dynamics}, we model the evolution of the translational dynamics as a function of its state $v$ and the control input $Rf_u$, where $f_u = \begin{bmatrix} 0&0&T
\end{bmatrix}^\mathrm{T}$, $T$ being the magnitude of the thrust generated by the quadrotor motors at the sampling time instant.

The quadrotor output wrench $\eta =  \begin{bmatrix} T&\tau_x&\tau_y&\tau_z
\end{bmatrix}^\mathrm{T}$ is linearly related to the squared motor speeds $u = \begin{bmatrix} n_1^2&n_2^2&n_3^2&n_4^2
\end{bmatrix}^\mathrm{T}$ by $\eta = B_0u$, where $B_0$ models the coefficients of rotor force and torque. Therefore, inspired by Eq~\eqref{eq:rotational_dynamics}, for our analysis, we will model the partial attitude dynamics for the quadrotor as follows:
\begin{equation}
    \dot{\omega} = g(\omega, u=\begin{bmatrix} n_1&n_2&n_3&n_4
\end{bmatrix}^\mathrm{T}).
\label{eq:attitude_partial_dynamics}
\end{equation}
Finally, based on Eq~\eqref{eq:partial_dynamics} and Eq~\eqref{eq:attitude_partial_dynamics}, we model the full system dynamics as follows:
\begin{equation}
\begin{bmatrix}
\dot{v} \\ 
\dot{\omega} 
\end{bmatrix}
= h\left(
\begin{bmatrix}
v \\ 
\omega 
\end{bmatrix}, u = \begin{bmatrix}
n_1^2 \\ 
n_2^2 \\
n_3^2 \\
n_4^2
\end{bmatrix}\right).
\label{eq:full_dynamics}
\end{equation}
We model a progression of partial and full dynamics to highlight the fact, as we will see in Sec~\ref{sec:Exp}, that LeARN becomes progressively more competitive against SINDy as the dimension of the state space increases. 

Now, our objective is to identify the mappings $f(\cdot)$ and $g(\cdot)$ and by extension the full dynamics $h(\cdot)$, as in Eq~\eqref{eq:partial_dynamics},~\eqref{eq:attitude_partial_dynamics} and~\eqref{eq:full_dynamics} respectively, without relying on a pre-defined library of basis functions (e.g., polynomial or trigonometric terms) as mandated by SINDy. 

\subsection{LeARN Framework}

Before introducing our framework, it is imperative to note that the greatest advantage of employing SINDy is its interpretability. For instance, an example dynamical evolution of a state vector $\textbf{x}\in\mathbb{R}^3$ could be learned via the SINDy framework and identified as follows:
\begin{equation}
    \dot{x} = 1.5*\sin(x[0]) + 0.25*\cos(x[1]),
\label{eq:sindy_example_evolution}
\end{equation}
where $sin(\cdot)$ and $cos(\cdot)$ comprise the candidate basis function library $\Theta(\cdot)$.
 Thus, any move away from SINDy will hamper its intepretability, as we see in ~\cite{c16}, which relies on latent coordinate transformations of system state vectors. Therefore, our attempt while designing the LeARN framework is to find a middle ground, as discussed next.

\begin{figure}[htp]
    \centering
    \includegraphics[width=\columnwidth]{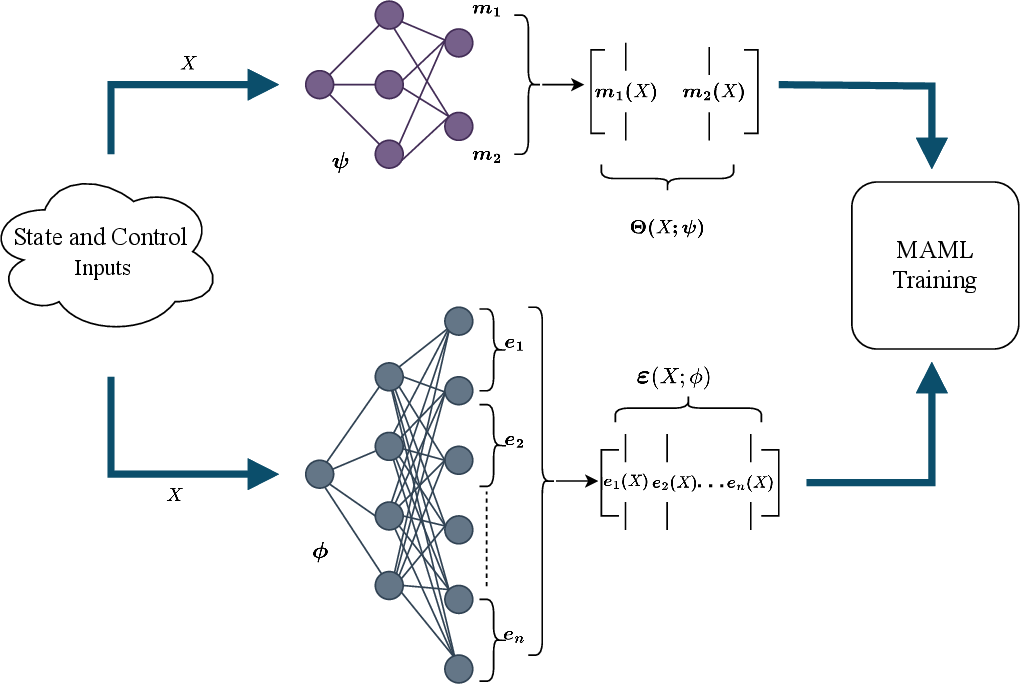}
    \caption{\small For a concatenated feature vector $X\in\mathbb{R}^{1 \times (I+U)}$ of the form, \( X = 
    \begin{bmatrix}
    x & u
    \end{bmatrix} \), where $x\in\mathbb{R}^{1 \times I}$ is the state feature vector and $u\in\mathbb{R}^{1 \times U}$ is the control input corresponding to $x$, $\Theta(X;\psi) \in \mathbb{R}^{1\times P(I+U)}$ is the
    learned basis function library for a total of P basis functions, where $m_p$ represents the $p^{th}$ basis function and $\mathcal{E}(X;\phi) \in \mathbb{R}^{I\times P(I+U)}$ is the learned feature selection matrix, for a total of $n = I$ sets of coefficients, $e_k \in \mathbb{R}^{1 \times (P+U)}$ is coefficient for $k^{th}$ state feature in $X$.}
    \label{fig:learn_framework}
\end{figure}

To avoid manual design (as in SINDy), we learn the basis function library directly from the data by parameterizing it via deep neural networks. We also seek to retain some key elements of the dynamical evolution, while employing the expressiveness of neural networks. Thus, our LeARN framework decomposes the dynamics into a learned basis function library and a feature selection matrix as shown in Fig~\ref{fig:learn_framework}. For an augmented input feature vector $X$ (combining state and control inputs), we model the dynamical evolution $y \approx \dot{\mathbf{x}}$ as
\begin{equation}
    y = \Theta(X;\psi) \mathcal{E}(X;\phi)^{T}.
\end{equation}

Here, $\Theta(\cdot;\psi)$ is a lightweight DNN parameterized by $\psi$ that learns a representation for the basis function library $\Theta(\cdot)$, and $\mathcal{E}(\cdot;\phi)$ is a second DNN parameterized by $\phi$ that acts as a \textit{feature selection matrix}. This decomposition structure mimics that of SINDy ($y = \Theta \xi$), where we can identify the terms contributing to the dynamics but replaces the fixed library with a learned parametrized representation of $\Theta$. Compared to Eq~\ref{eq:sindy_example_evolution}, LeARN could estimate the dynamical evolution of some state vector $\textbf{x}\in\mathbb{R}^3$ as follows:
\begin{equation}
    \dot{x} = 1.64*m_1(x[0]) + 0.24*m_2(x[1]),
\label{eq:learn_example_evolution}
\end{equation}
where $m_1(\cdot)$ and $m_2(\cdot)$ are the outputs of the DNN $\Theta(\cdot;\psi)$ as illustrated in Fig~\ref{fig:learn_framework}. In Eq~\ref{eq:sindy_example_evolution}, the functional relationship and the dominant terms that contribute to the dynamics are apparent. LeARN retains the latter, from Eq~\ref{eq:learn_example_evolution}, we infer that $x[0]$ and $x[1]$ contribute predominantly to the dynamical evolution of $\textbf{x}\in\mathbb{R}^3$, while $x[2]$ does not contribute to its evolution. Note that while each $m_p(\cdot)$ takes the full feature vector 
$X$ as input (see Fig.~\ref{fig:learn_framework}), interpretability 
here refers to the sparsity structure of $\mathcal{E}$, which 
identifies dominant input features, rather than to the functional 
form of $m_p$ itself.

\subsection{Meta-Training and Adaptation}
We start training by employing a Model-Agnostic Meta-Learning (MAML) approach to learn $\psi$ and $\phi$. The training loop samples tasks $\mathcal{T}_i$ corresponding to different wind conditions from the training distribution. Since we are evaluating our methods
on the Neural Fly~\cite{c3} dataset, the different task objectives involve modeling the
system dynamics accurately for a quadrotor being flown using a nonlinear baseline controller under
different wind conditions. In the inner loop, the parameters are adapted to minimize the dynamical estimation loss $L_i = \| \dot{v}_i - \Theta_\psi \mathcal{E}_{\phi}^T \|_2^2$. The outer loop updates the initial parameters to ensure that they serve as a robust initialization for rapid adaptation to unseen environments. The algorithm is presented in a concise manner in Alg.~\ref{alg:meta-learning}.

\begin{algorithm}[t]
\caption{The Meta-Training Loop of LeARN}
\label{alg:meta-learning}
\SetNoFillComment
\KwIn{[$v(W), Rfu(W)$]: feature distributions over $\mathcal{D}_i^{train}$}
\KwOut{$\phi,\psi$: meta-trained parameters}
\SetKwInOut{Hyperparameters}{Hyperparameters}
\Hyperparameters{$\alpha, \beta, n$ (step sizes, inner rounds)}

\tcc{Randomly initialize $\phi$ and $\psi$ (parameters of the DNN giving $\mathcal{E}$ and basis library functions in $\Theta$)}

\While{not done}{
    Sample batch of tasks $\mathcal{T}_i(v_i, Rfu_i) \sim [v(W), Rfu(W)]$\;
    
    \For{all $\mathcal{T}_i(v_i, Rfu_i)$}{
        Evaluate basis $\Theta_\psi(v_i, Rfu_i)$ and selection $\mathcal{E}_{\phi}(v_i, Rfu_i)$\;
        
        \tcp*[l]{Evaluate the task objective (dynamical estimation) loss:}
        $L_i \leftarrow \left\| \dot{v}_i - \Theta_\psi(v_i, Rfu_i) \mathcal{E}_{\phi}(v_i, Rfu_i) \right\|_2^2$\;

        \For{$\text{round} \leftarrow 1$ \KwTo $n$}{
            \tcp*[l]{Compute adapted parameters $\phi'$ and $\psi'$:}
            $\phi' \leftarrow \phi - \alpha \cdot \nabla_\phi L_i$\;
            $\psi' \leftarrow \psi - \alpha \cdot \nabla_\psi L_i (\text{using } f_\psi)$\;
        }
        Sample query points $D_{i}' = \{v_i', Rfu_i'\}$ from $\mathcal{T}_i$\;
    }
    
    \tcp*[l]{Consider meta-loss $L_i'$ and perform meta-update:}
    $L_i' \leftarrow \left\| \dot{v}_i' - \Theta_{\psi'}(v_i', Rfu_i') \mathcal{E}_{\phi'}(v_i', Rfu_i') \right\|_2^2$\;
    $\phi \leftarrow \phi - \beta \cdot \nabla_\phi \sum_{D_i'} L_i'$\;
    $\psi \leftarrow \psi - \beta \cdot \nabla_\psi \sum_{D_i'} L_i'$\;
}
\end{algorithm}
Once, we have the meta-trained parameters, we perform online adaptation while deployment on unseen new tasks as discussed in Eq~\eqref{eq:online_adapt}. While online adaptation, we assume that the state features are $L$ Lipschitz continuous, and add a regularizer to guide the training process as follows:
\begin{align} \label{eq:online_adapt}
    \psi_{\text{t+1}} &= \psi_{\textit{t}} - \nabla_{\psi}L_{\textit{adapt}}(f(X_{\textit{t}};\phi_{\textit{t}},\psi_{\textit{t}})),  \\ 
    \phi_{\text{t+1}} &= \phi_{\textit{t}} - \nabla_{\phi}L_{\textit{adapt}}(f(X_{\textit{t}};\phi_{\textit{t}},\psi_{\textit{t}})),   
\end{align}
where we initialize $\psi_{\textit{t}}$ and $\phi_{\textit{t}}$, with the meta-trained parameters $\psi$ and $\phi$ respectively at $t=0$, while adapting to a new task i.e. a new wind condition for the quadrotor to fly in, $X_{\textit{t}}$ is sampled from the new task and $L_{\textit{adapt}}$ gives us the adaptation loss as follows:
\begin{align} \label{eq:lipschitz}
L_{\text{adapt}}(f_t) &= \left( \dot{v}_t - f(X_t; \phi_t, \psi_t) \right)^2 + \lambda \max \bigl( 0, \\
& \|f(X_t; \phi_t, \psi_t) - f(X_{t-1}; \phi_{t-1}, \psi_{t-1})\|_1 - L \bigr),
\end{align}
where $\lambda$ is a hyperparameter and the regularizer ensures $L$ Lipschitz continuity of the state inputs.

\section{Experiments and Results}
\label{sec:Exp}

\subsection{Experimental Setup}

We present a comparative performance analysis of our framework LeARN against SINDy. Based on the Neural Fly dataset~\cite{c3}, we use data gathered by the quadrotor using the baseline controller in the wind conditions as documented in Table~\ref{tab:wind_conditions} for meta-training the parameters $\psi$ and $\phi$.
\begin{table}[h]
\centering
\caption{Wind Conditions for Meta-Training}
\label{tab:wind_conditions}
\begin{tabular}{ll | ll}
\hline
\textbf{Condition} & \textbf{Speed} & \textbf{Condition} & \textbf{Speed} \\ \hline
nowind  & 0.0 m/s & 30wind  & 3.7 m/s \\
10wind  & 1.3 m/s & 40wind  & 4.9 m/s \\
20wind  & 2.5 m/s & 50wind  & 6.1 m/s \\ \hline
\end{tabular}
\end{table}
We evaluate LeARN and SINDy for the system identification tasks across three different formulations of the quadrotor dynamics,the partial translational dynamics formulated in Eq~\eqref{eq:partial_dynamics}, the attitude dynamics in Eq~\eqref{eq:attitude_partial_dynamics} and the full system dynamics in Eq~\eqref{eq:full_dynamics}. We perform this evaluation for the data gathered by the baseline nonlinear controller in the following wind conditions as documented in Table~\ref{tab:evaluation_wind}.
\begin{table}[h]
\centering
\caption{Wind Conditions for Evaluation}
\label{tab:evaluation_wind}
\begin{tabular}{ll | ll}
\hline
\textbf{Condition} & \textbf{Speed} & \textbf{Condition} & \textbf{Speed} \\ \hline
35wind  & 4.2 m/s & 70psin20 & $8.5 + 2.4 \sin(t)$ m/s \\
70wind  & 8.5 m/s & 100wind  & 12.1 m/s \\ \hline
\end{tabular}
\end{table}
We use a lightweight, fully connected DNN parameterized by $\psi$, with GELU activations to model the basis function library $\Theta$. This DNN has the following layer dimensions: 
\begin{align*}
     [(&\textit{Input Dim.} \times 12), (12 \times 16), \\
    &(16 \times 24), (24 \times 48), \\&(48 \times \textit{\# Outputs for 
    basis function representation})].
\end{align*}
We also use another lightweight fully connected DNN parameterized by $\phi$, with GELU activations to model the feature selection matrix $\mathcal{E}$. To evaluate against the right set of basis functions in SINDy, from extensive experimentation, we have seen that for the Neural Fly dataset, $sin(x)$ and $cos(x)$ suffice for constructing the basis function library for the SINDy framework. Therefore, for competitive comparison with LeARN, we model the DNN to have only two output features as illustrated in Fig~\ref{fig:learn_framework}, which gives us a set of two learned analogues of basis functions. 

We employ SINDy with the sparsity threshold set to 0.2. Before delving into further results, it is important to note that for the adaptation wind tasks in the Neural Fly dataset, the data was collected by maneuvering the quadrotor in a figure-eight trajectory. The figure-eight trajectory is inherently oscillatory, which makes it naturally suited to modeling by trigonometric functions. Although DNNs are universal approximators, their capacity is constrained by their architecture and training data. A finite-capacity DNN may have insufficient expressive power 
to capture the intricate oscillatory behavior of such trajectories, 
particularly in unobserved regions of the input space, resulting 
in suboptimal generalization performance.
 

%

\begin{figure*}[p]
    \centering
    \renewcommand{\arraystretch}{1.5}
    \setlength{\tabcolsep}{1pt}
    \begin{tabular}{c|c}
        \multicolumn{2}{r}{%
            {\setlength{\fboxsep}{4pt}%
            \fbox{\small
                \textcolor[HTML]{FFAA6B}{\rule{12pt}{5pt}}~Ground Truth\quad
                \textcolor[HTML]{EB3324}{\rule{12pt}{5pt}}~Adaptation\quad
                \textcolor[HTML]{3282F6}{\rule{12pt}{5pt}}~Generalization%
            }}%
        }\\[-4pt]
        \small SINDy (35wind) & \small LeARN (35wind) \\ \hline
        \begin{tabular}{ccc}
            \stackunder{\includegraphics[width=0.16\textwidth]{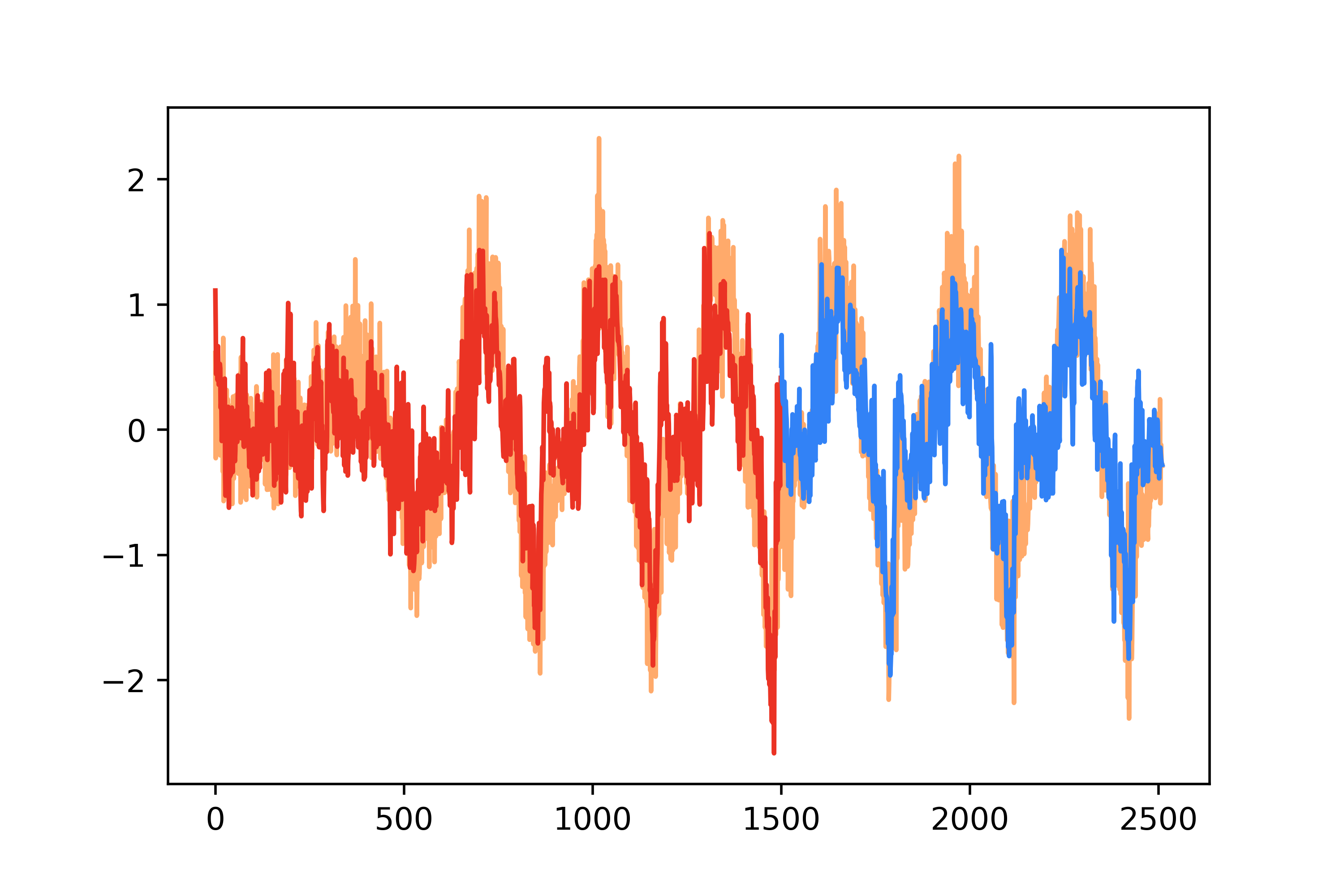}}{\scriptsize $\dot{v_x}$} &
            \stackunder{\includegraphics[width=0.16\textwidth]{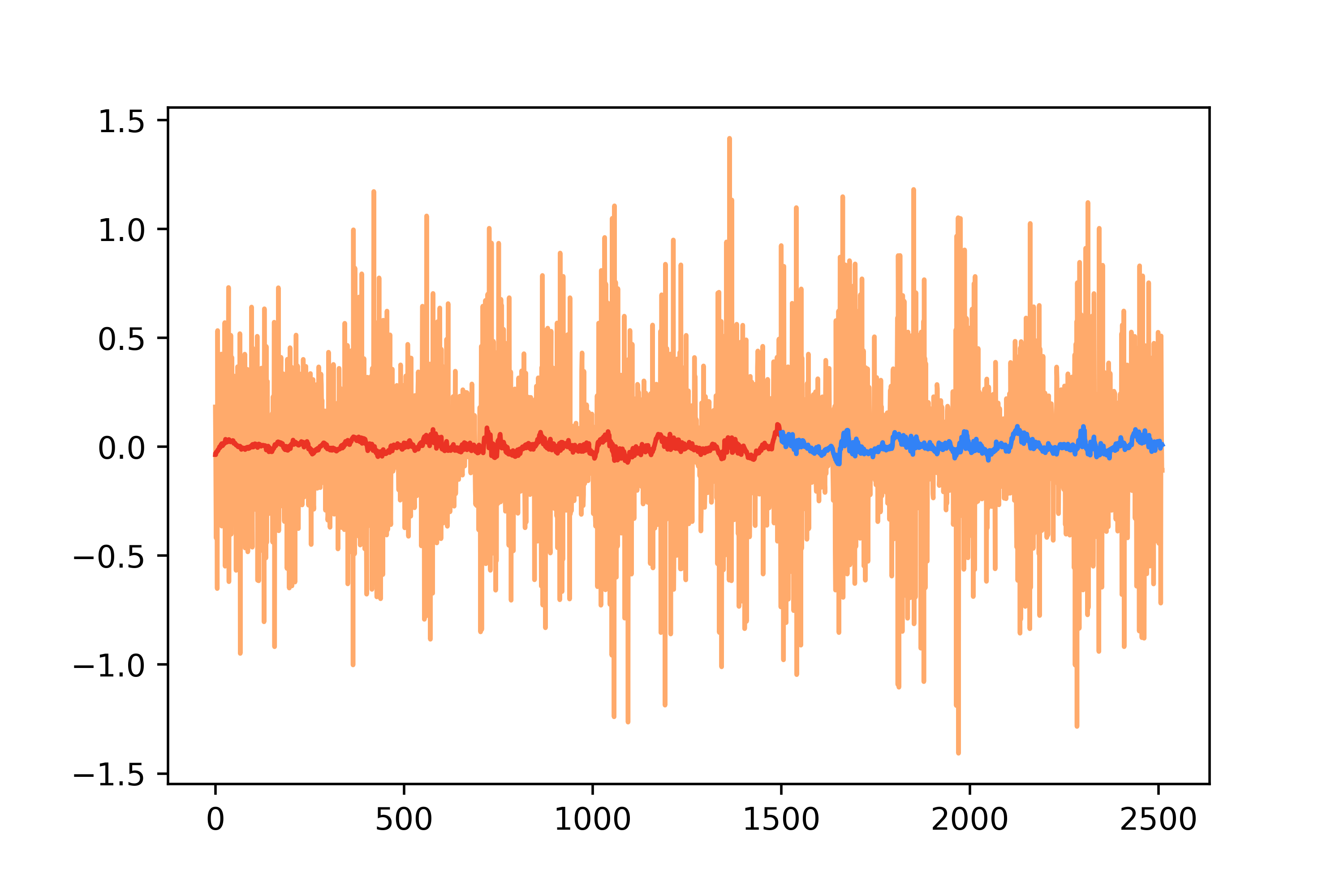}}{\scriptsize $\dot{v_y}$} &
            \stackunder{\includegraphics[width=0.16\textwidth]{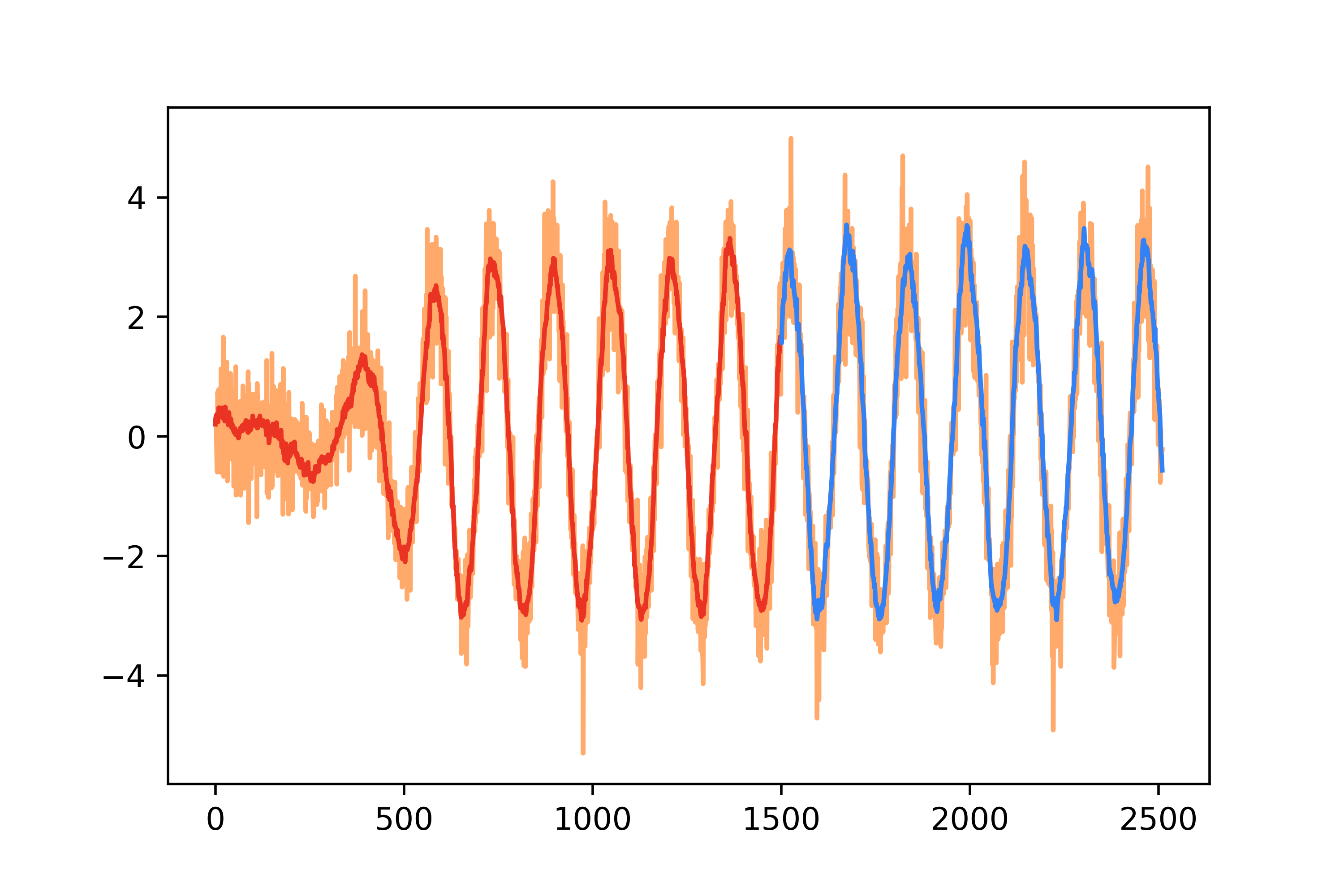}}{\scriptsize $\dot{v_z}$}
        \end{tabular} &
        \begin{tabular}{ccc}
            \stackunder{\includegraphics[width=0.16\textwidth]{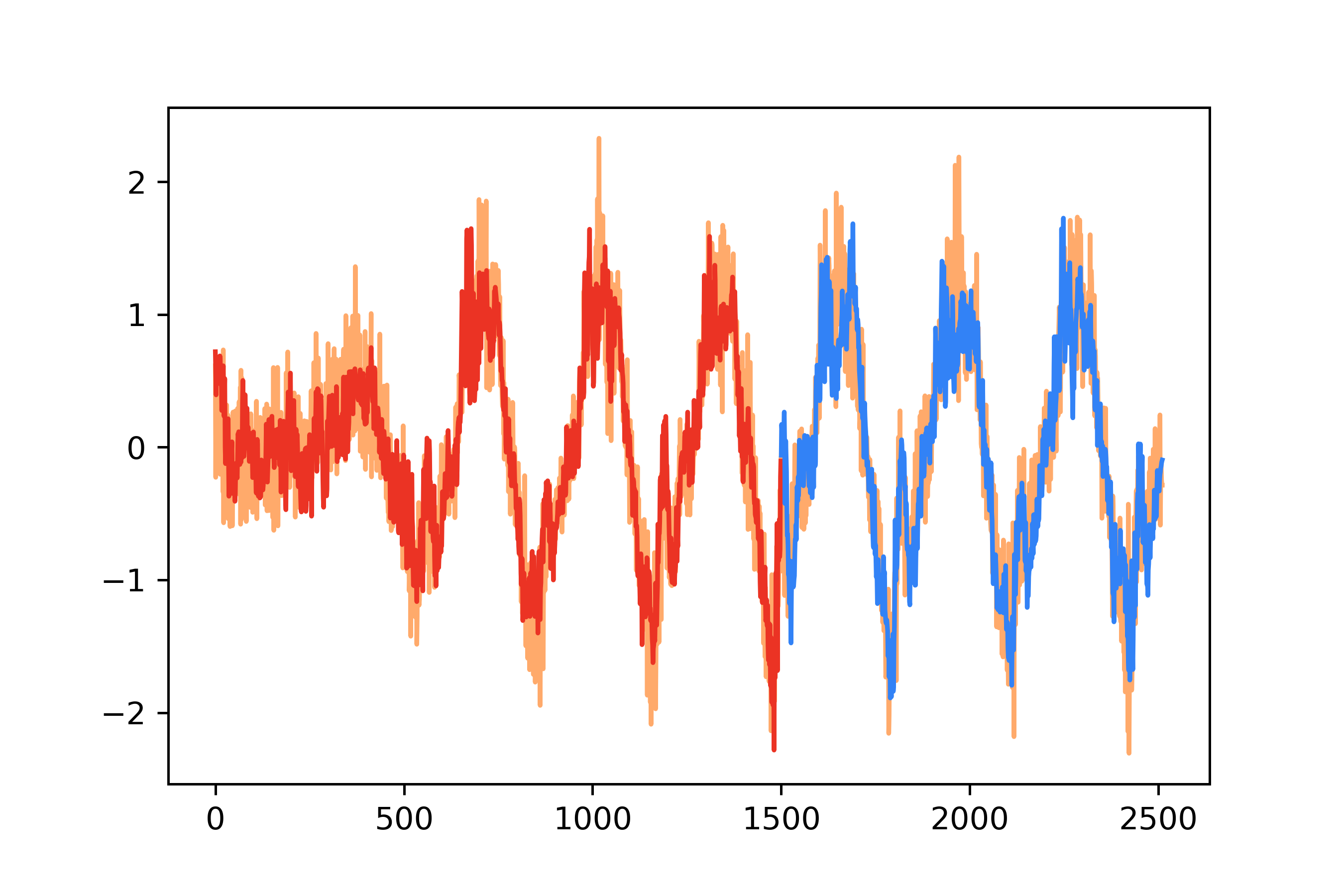}}{\scriptsize $\dot{v_x}$} &
            \stackunder{\includegraphics[width=0.16\textwidth]{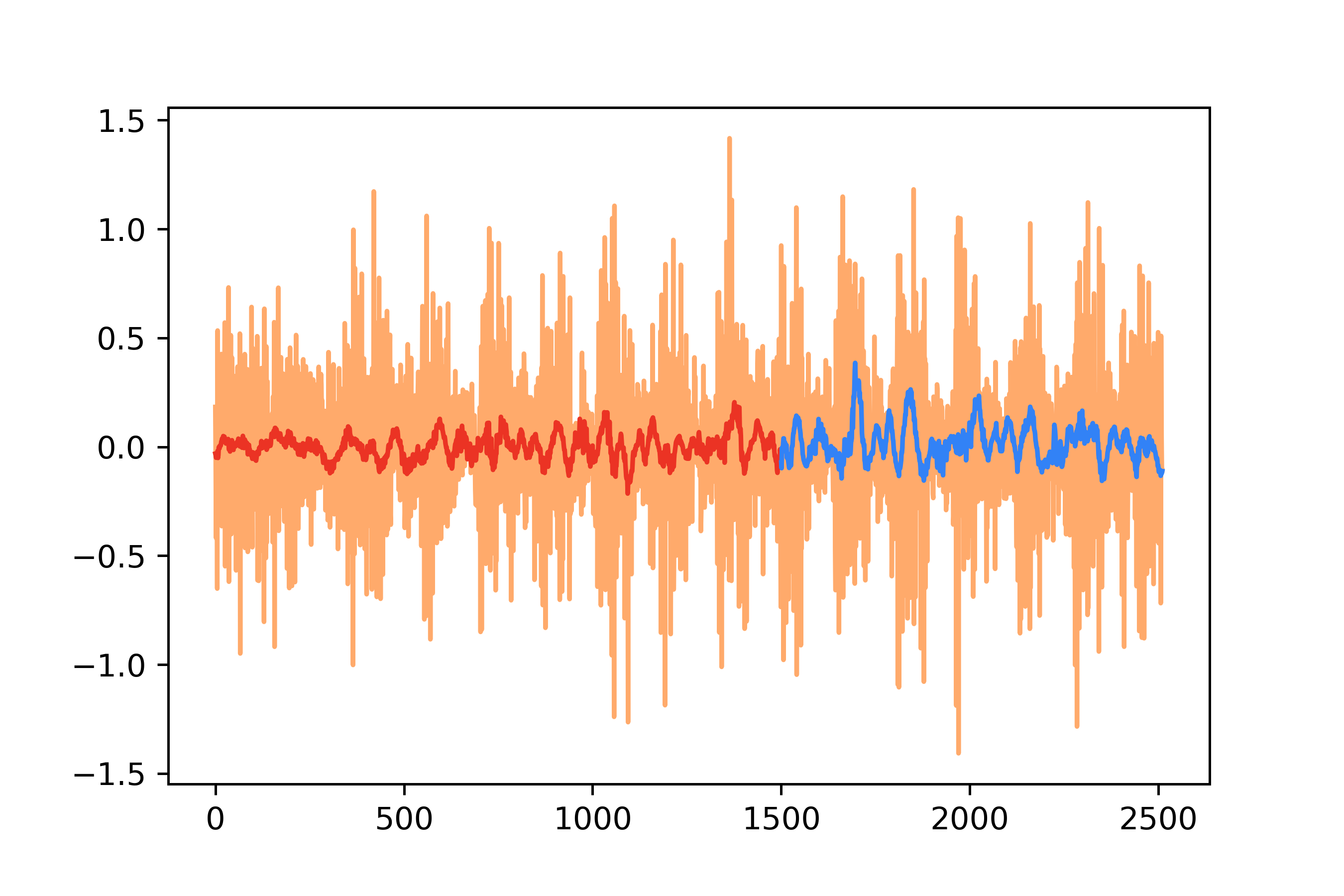}}{\scriptsize $\dot{v_y}$} &
            \stackunder{\includegraphics[width=0.16\textwidth]{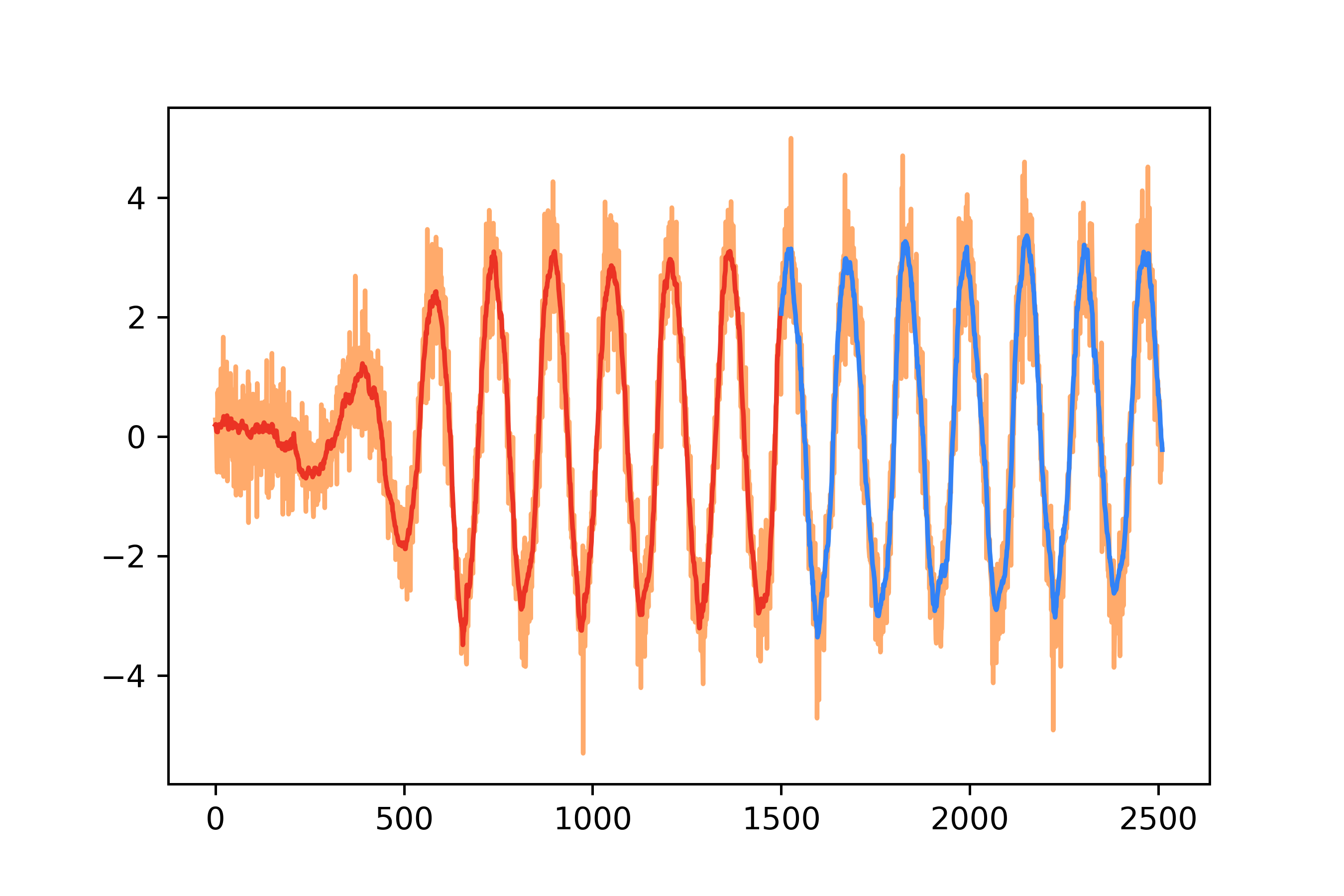}}{\scriptsize $\dot{v_z}$}
        \end{tabular} \\ \hline
        \begin{tabular}{ccc}
            \stackunder{\includegraphics[width=0.16\textwidth]{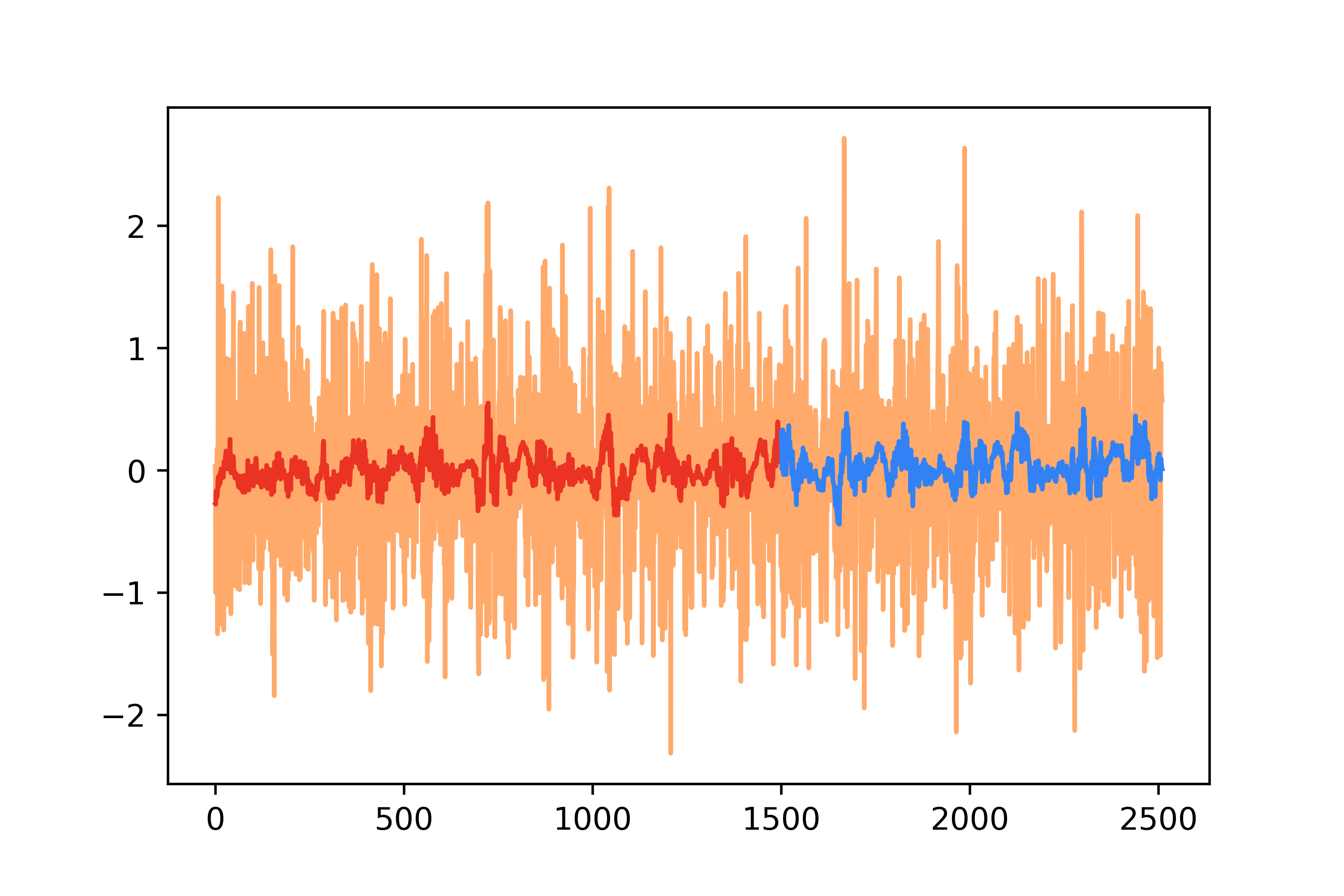}}{\scriptsize $\dot{\omega_x}$} &
            \stackunder{\includegraphics[width=0.16\textwidth]{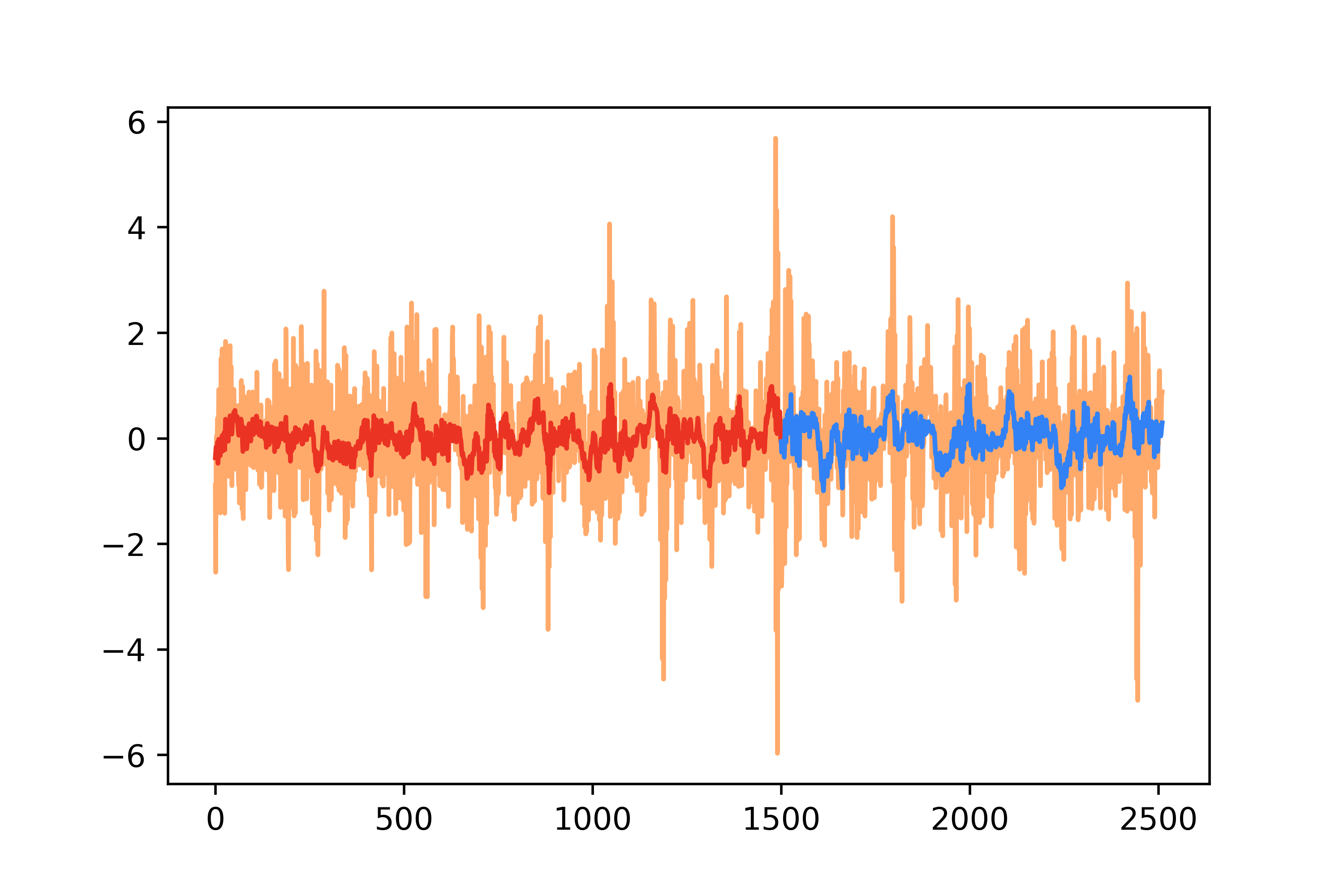}}{\scriptsize $\dot{\omega_y}$} &
            \stackunder{\includegraphics[width=0.16\textwidth]{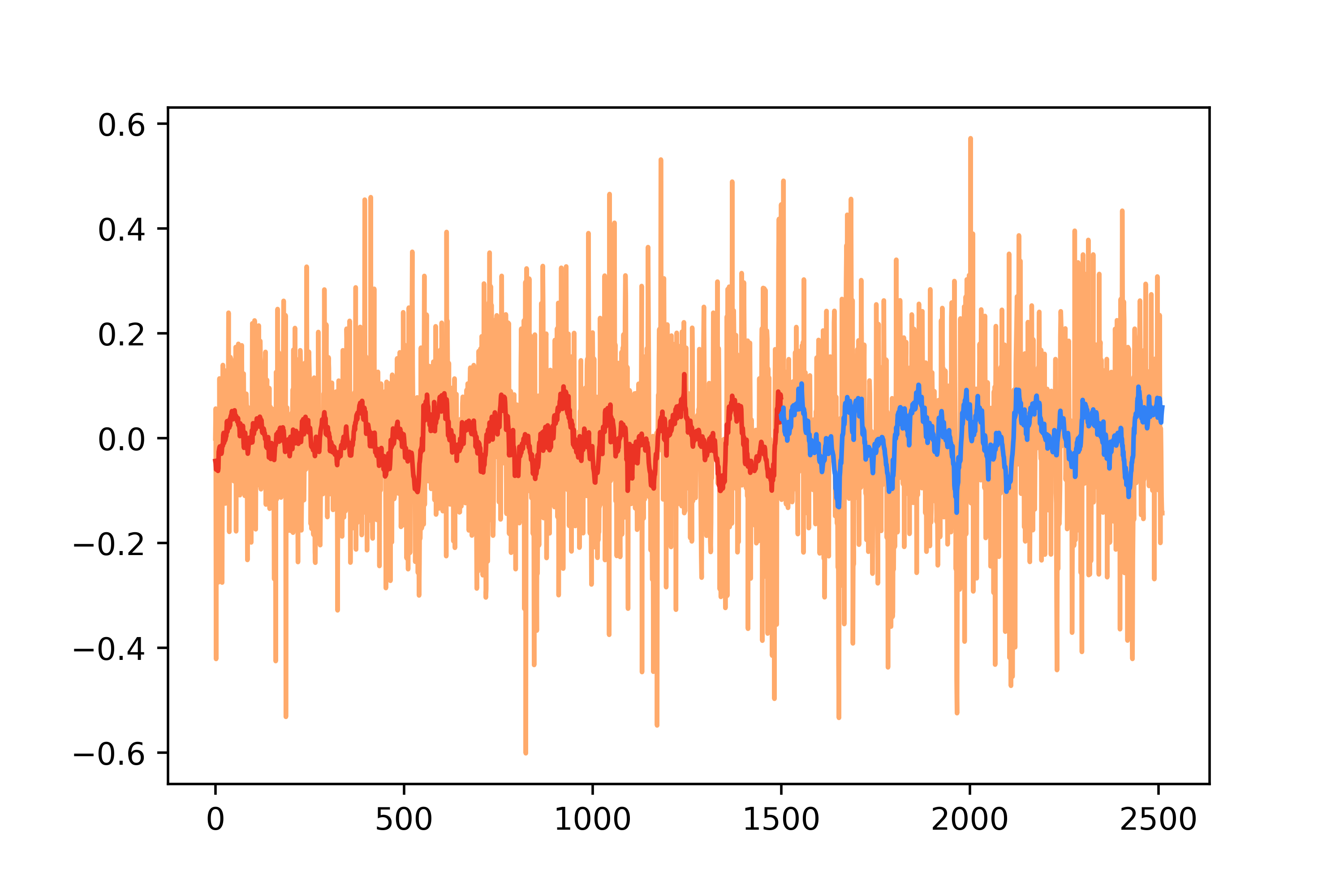}}{\scriptsize $\dot{\omega_z}$}
        \end{tabular} &
        \begin{tabular}{ccc}
            \stackunder{\includegraphics[width=0.16\textwidth]{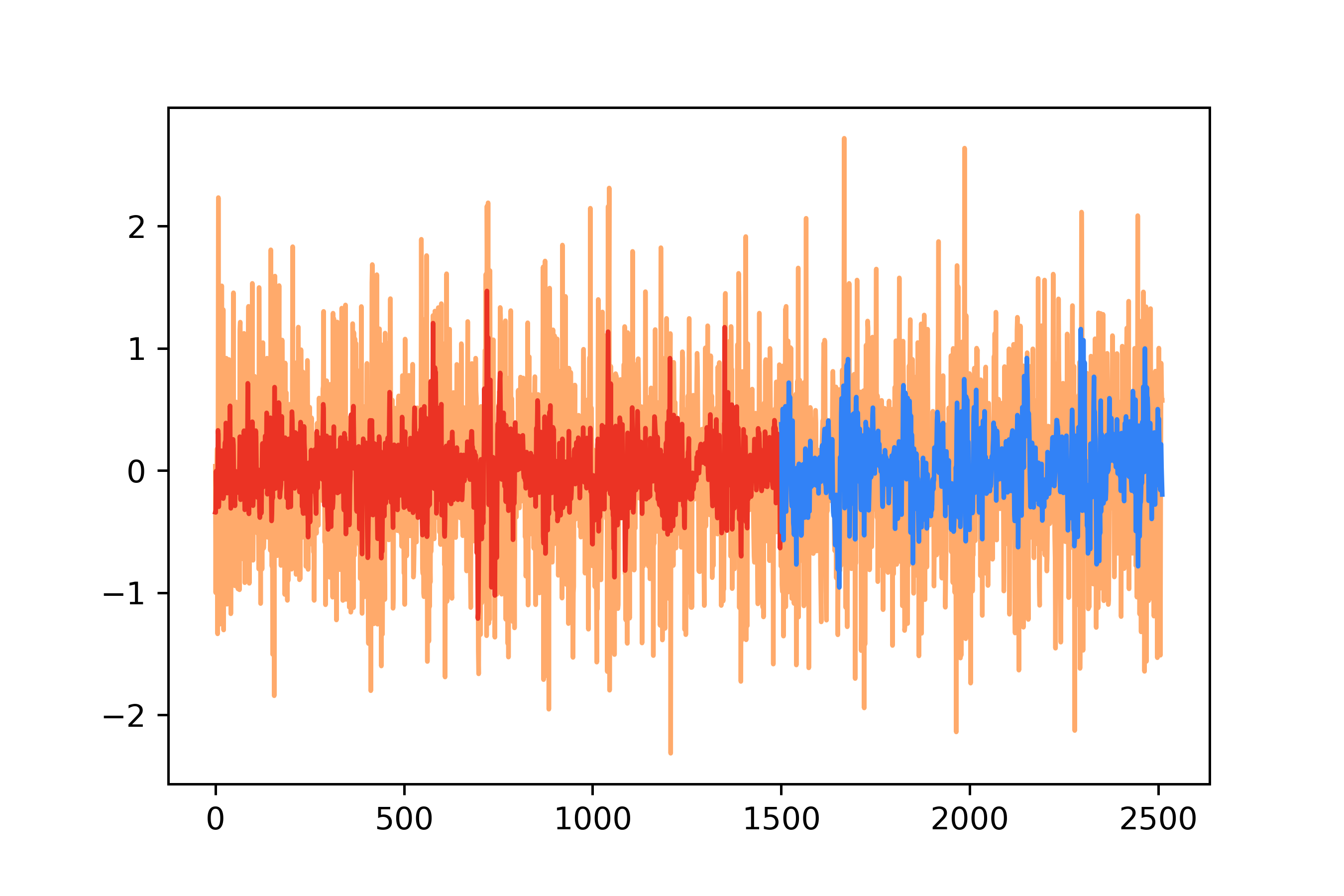}}{\scriptsize $\dot{\omega_x}$} &
            \stackunder{\includegraphics[width=0.16\textwidth]{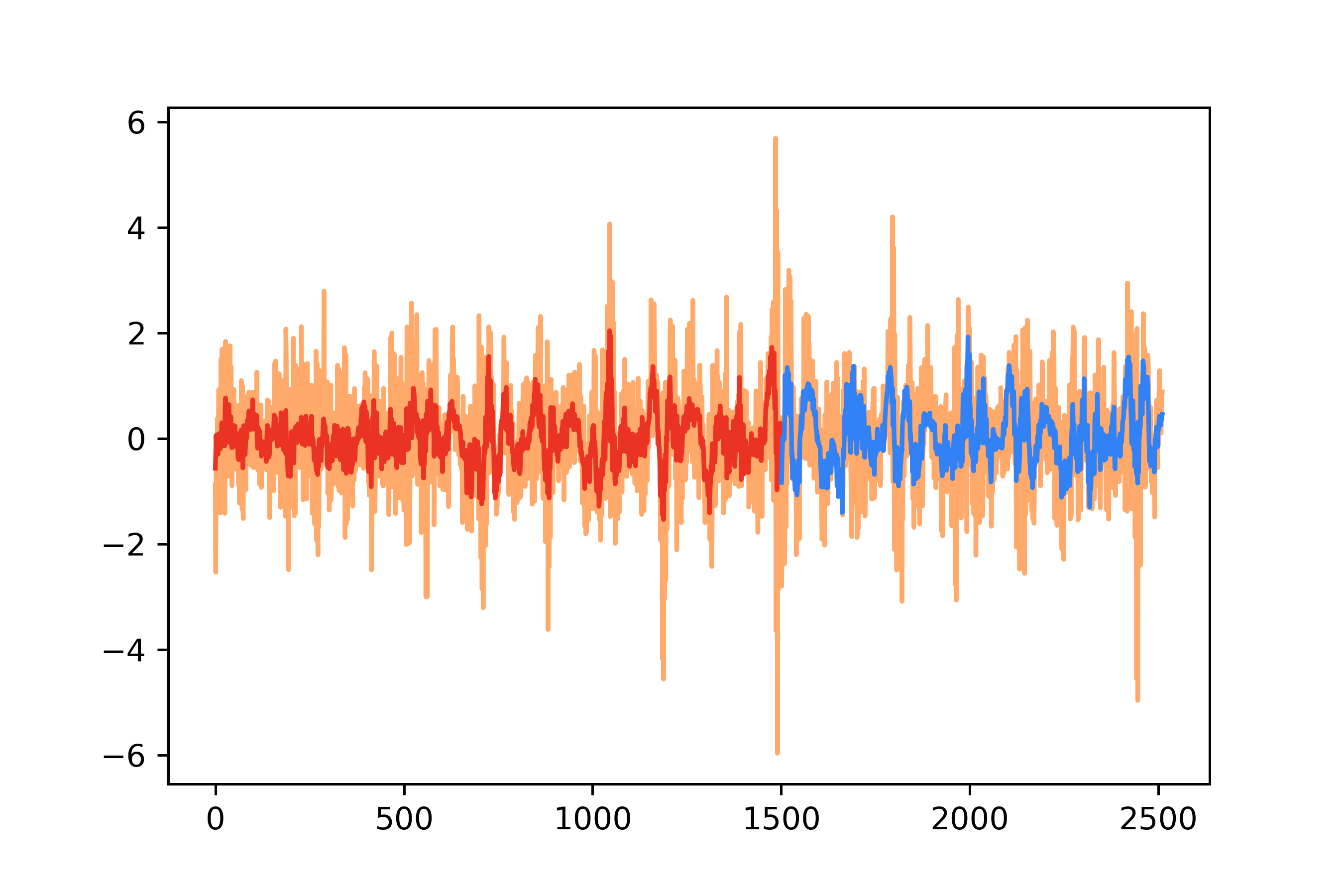}}{\scriptsize $\dot{\omega_y}$} &
            \stackunder{\includegraphics[width=0.16\textwidth]{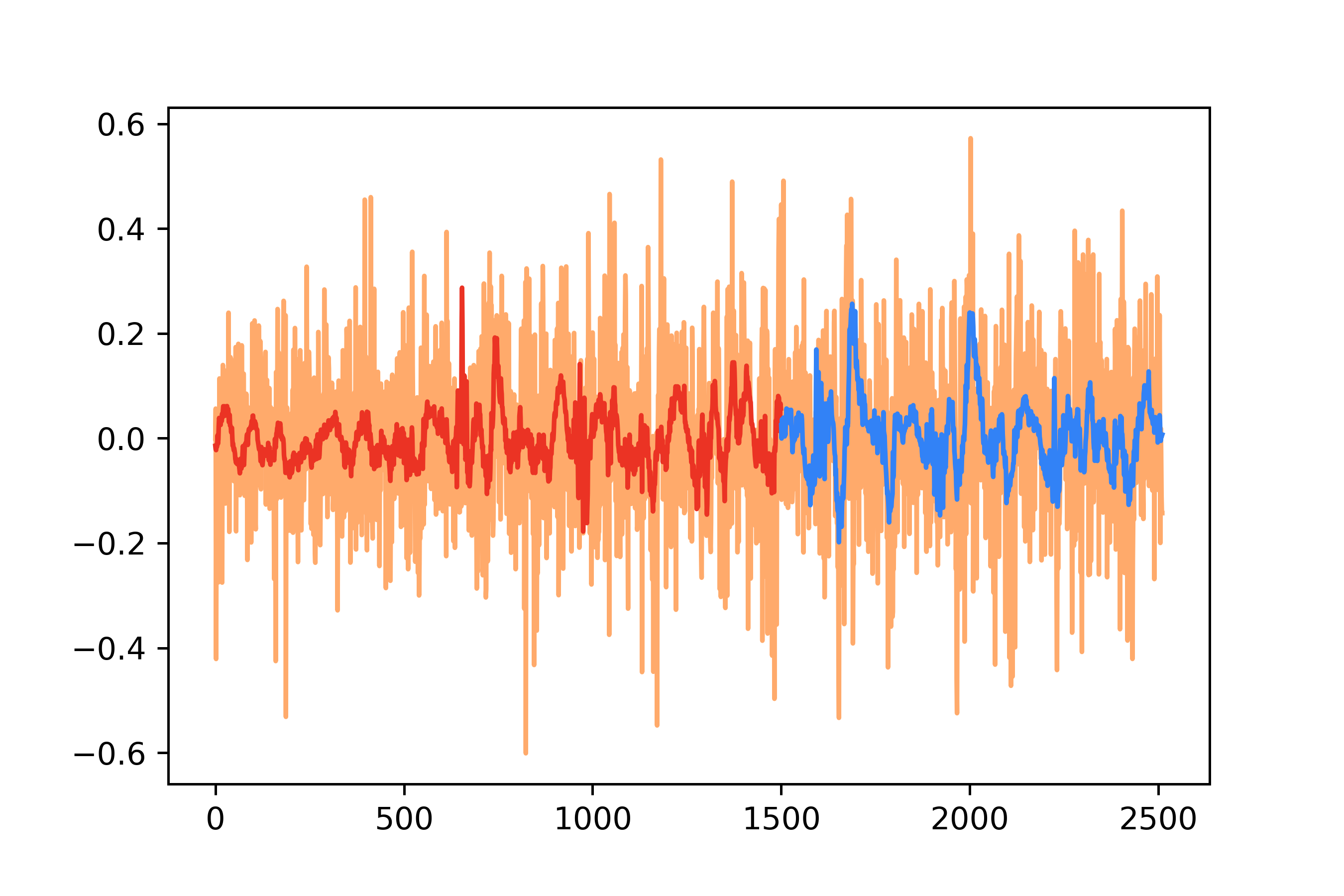}}{\scriptsize $\dot{\omega_z}$}
        \end{tabular} \\ \hline
        \small SINDy (70wind) & \small LeARN (70wind) \\ \hline
        \begin{tabular}{ccc}
            \stackunder{\includegraphics[width=0.16\textwidth]{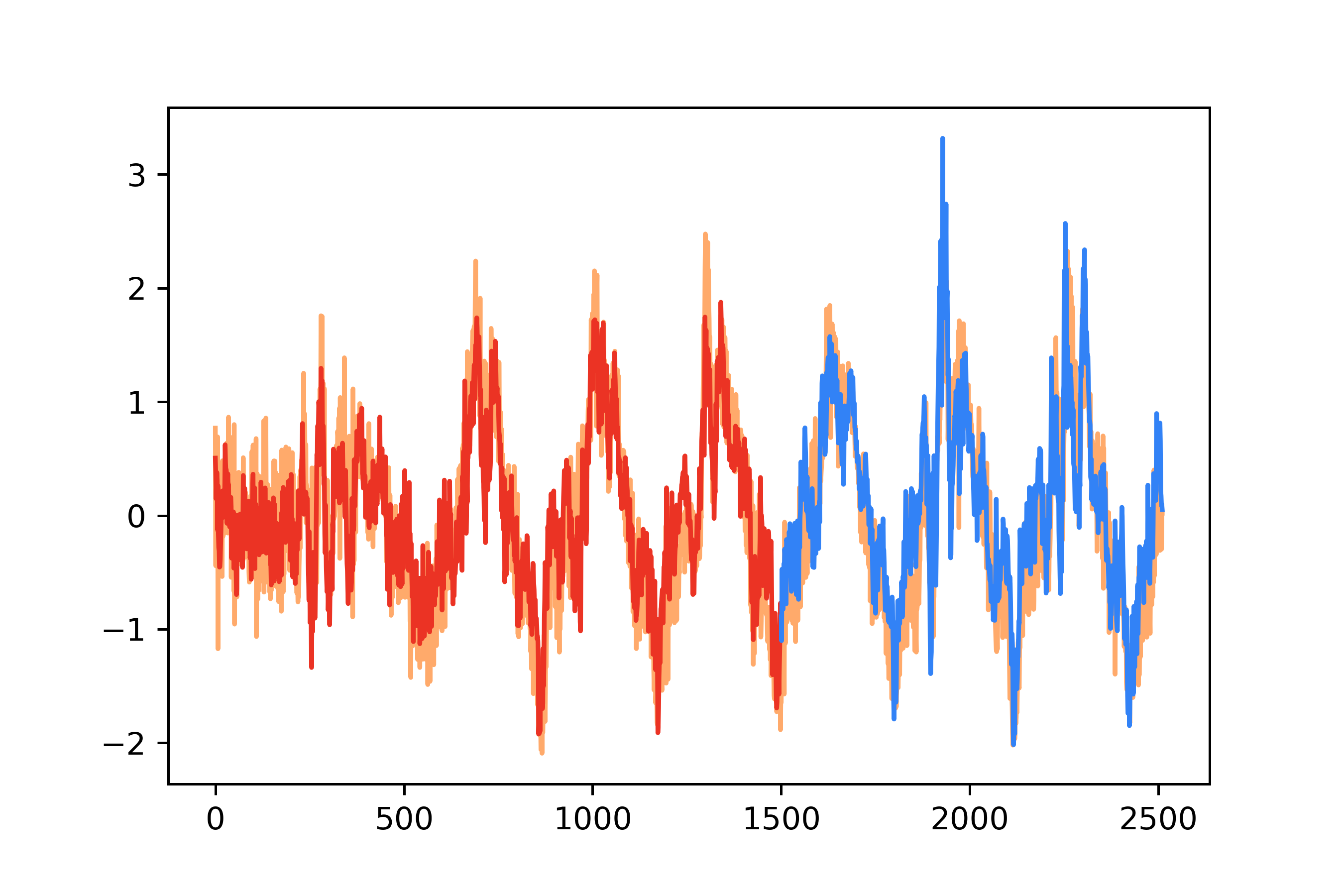}}{\scriptsize $\dot{v_x}$} &
            \stackunder{\includegraphics[width=0.16\textwidth]{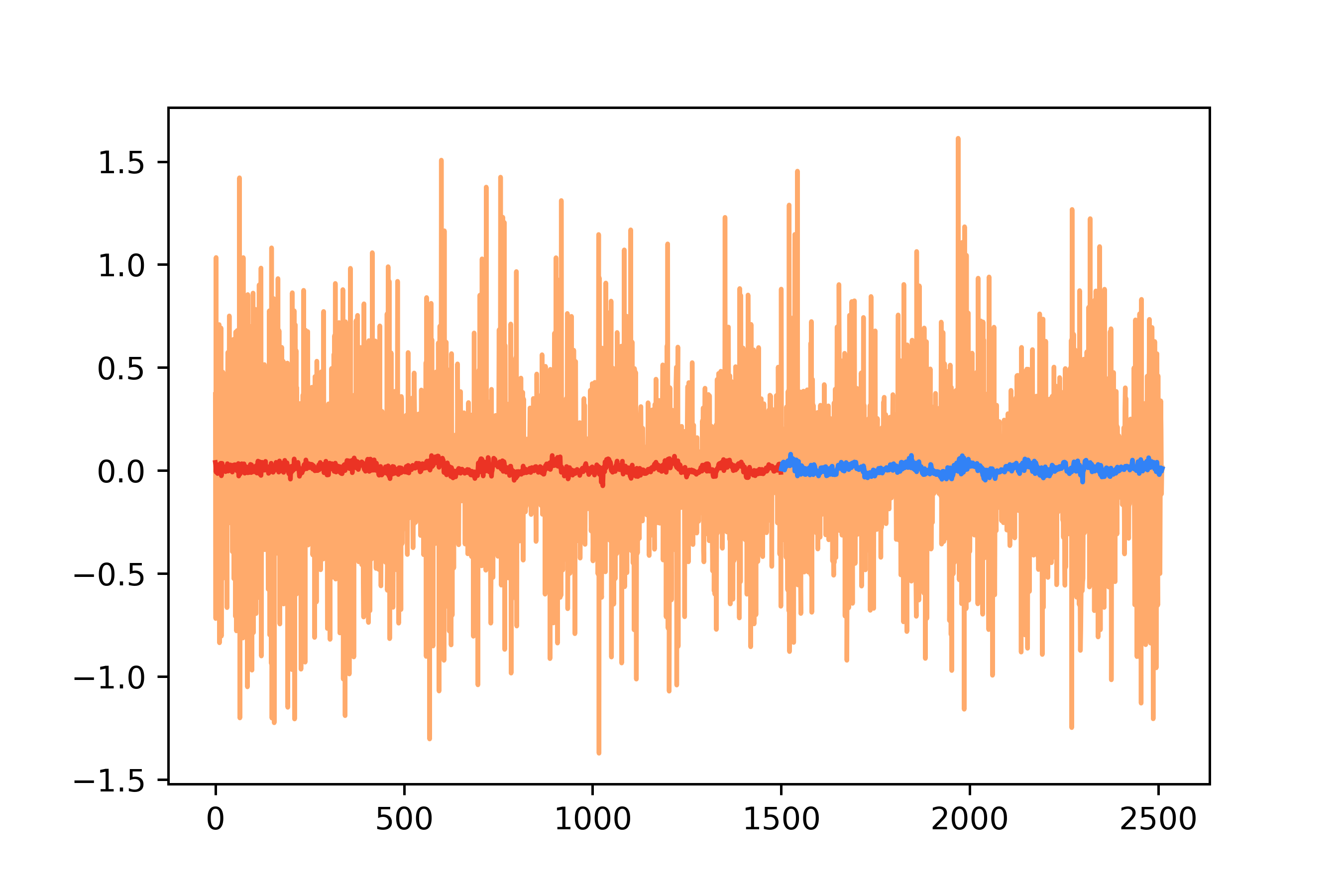}}{\scriptsize $\dot{v_y}$} &
            \stackunder{\includegraphics[width=0.16\textwidth]{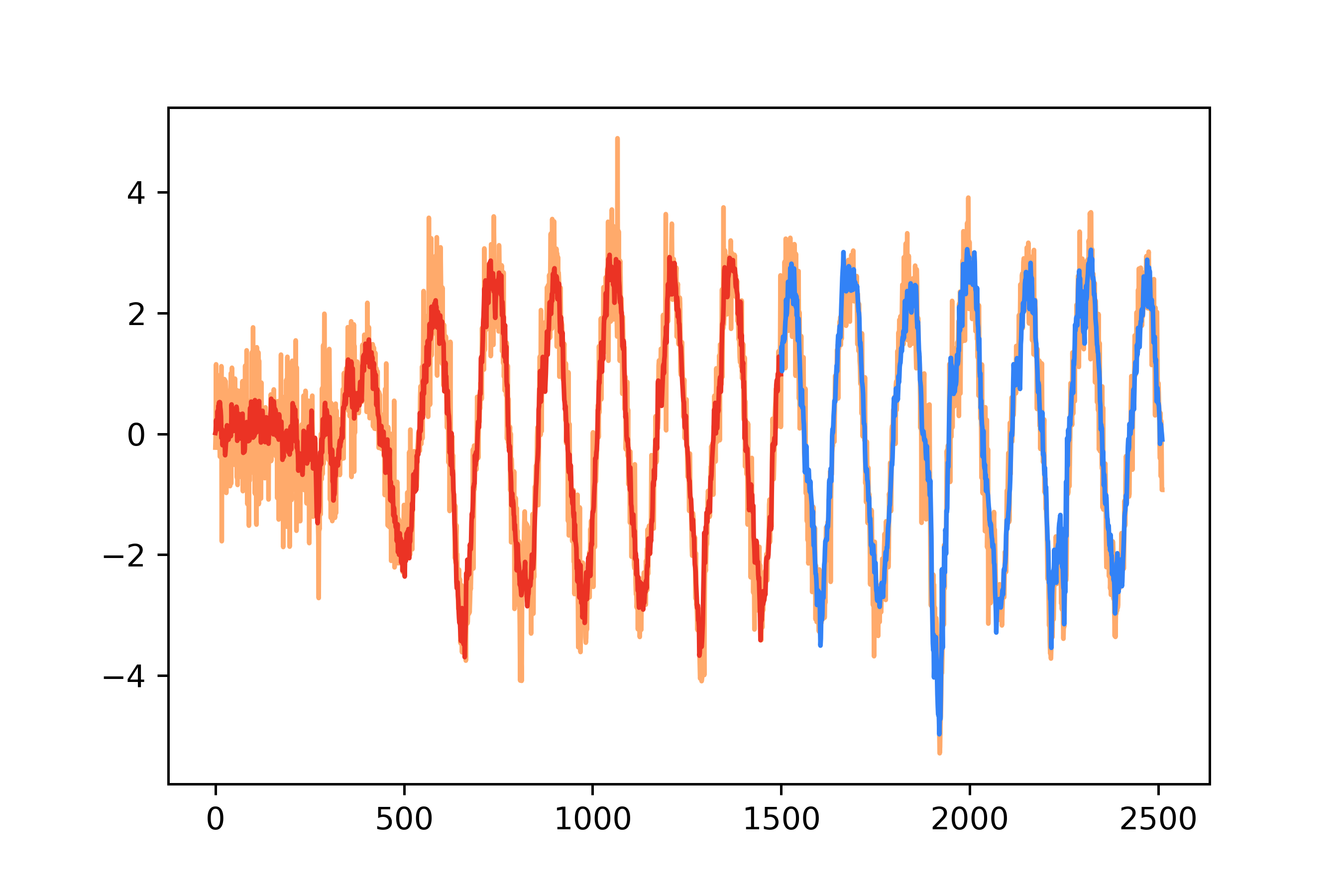}}{\scriptsize $\dot{v_z}$}
        \end{tabular} &
        \begin{tabular}{ccc}
            \stackunder{\includegraphics[width=0.16\textwidth]{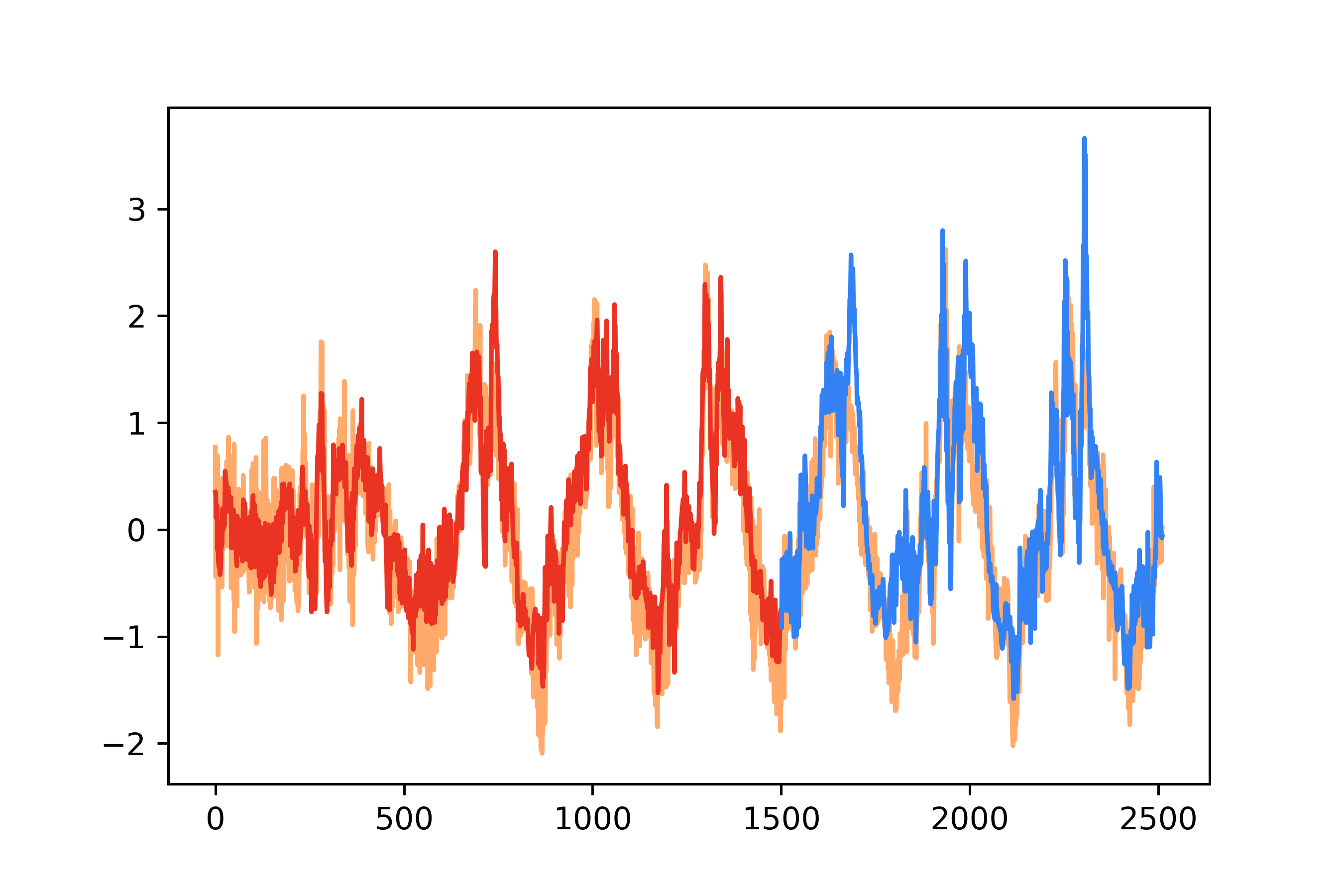}}{\scriptsize $\dot{v_x}$} &
            \stackunder{\includegraphics[width=0.16\textwidth]{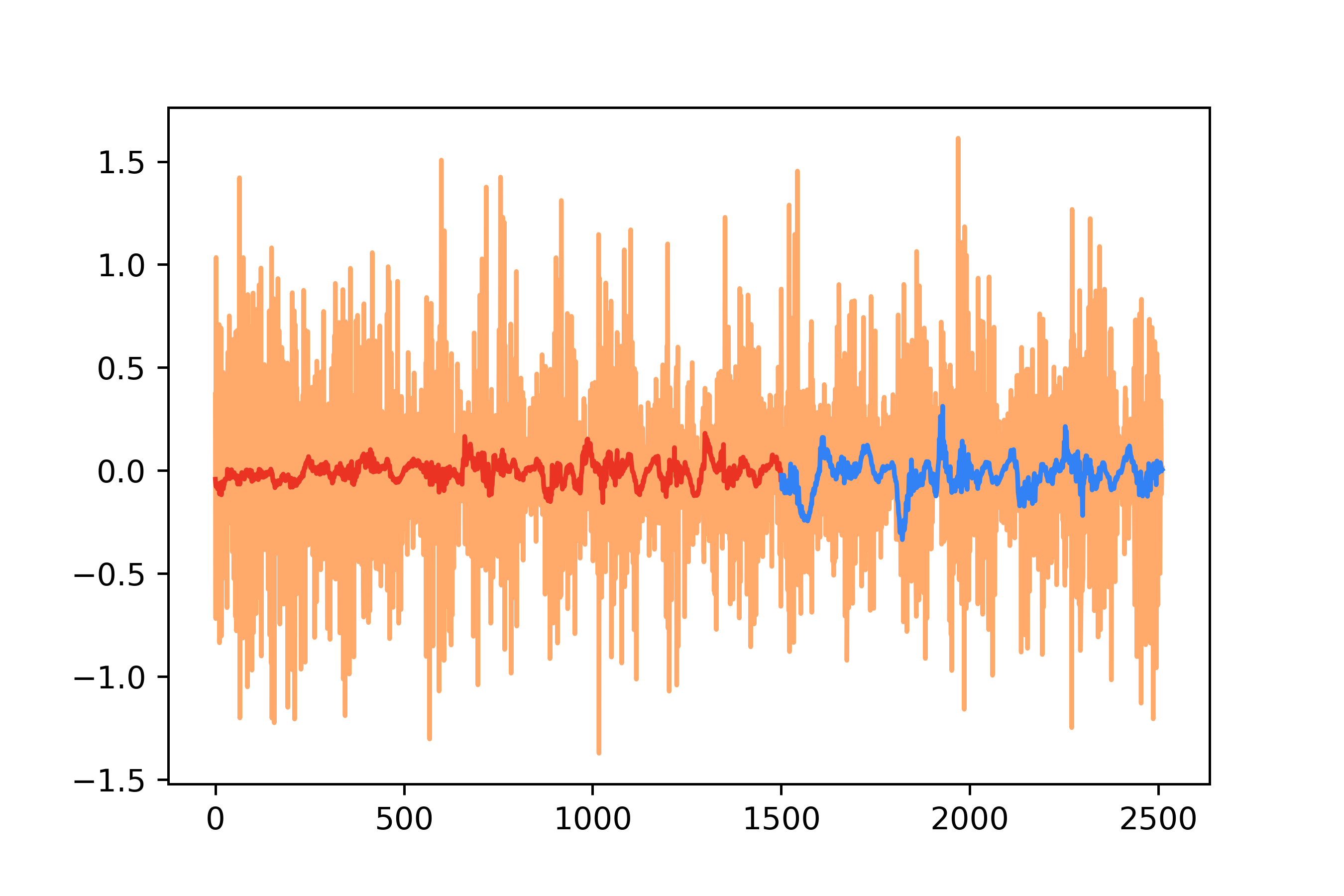}}{\scriptsize $\dot{v_y}$} &
            \stackunder{\includegraphics[width=0.16\textwidth]{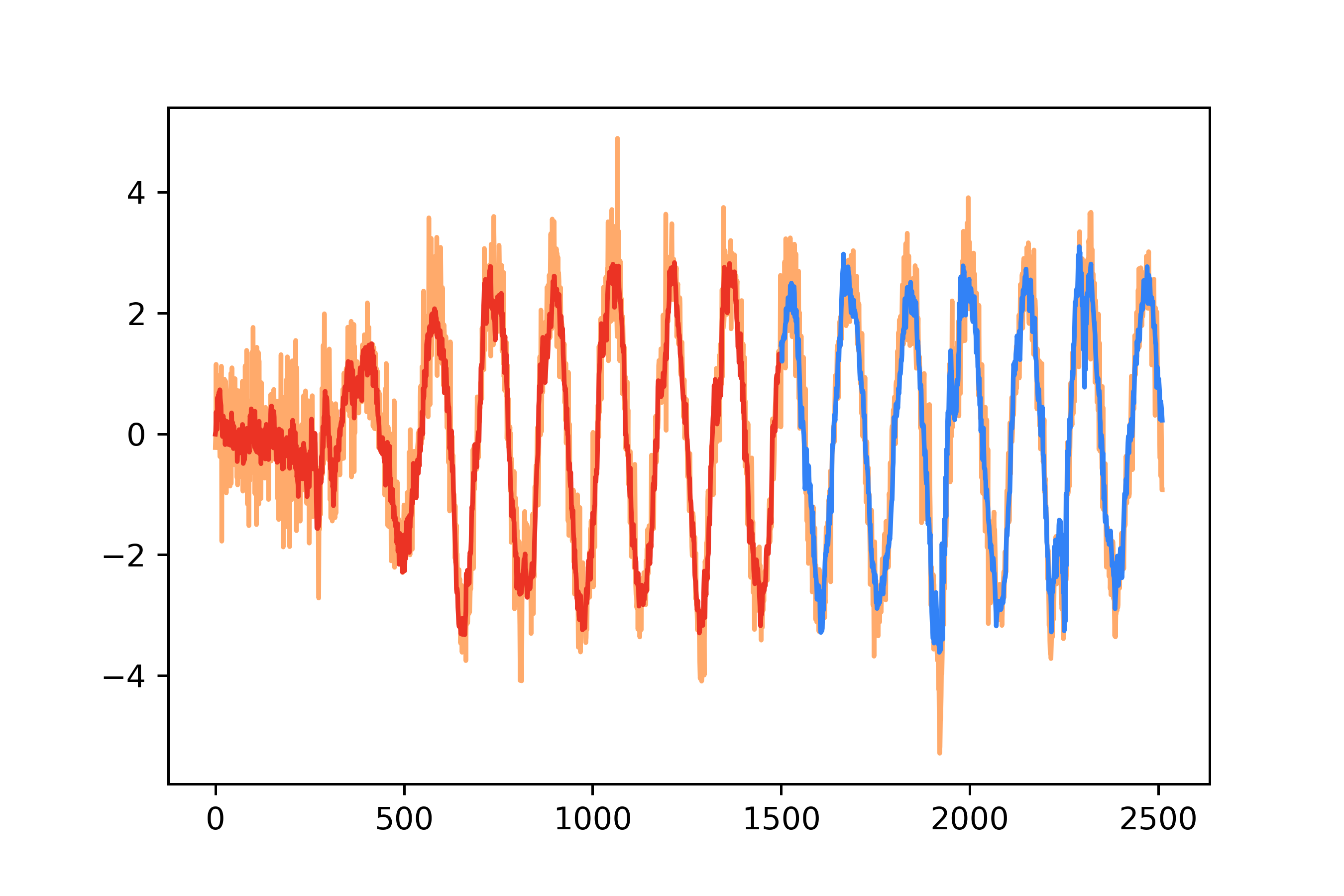}}{\scriptsize $\dot{v_z}$}
        \end{tabular} \\ \hline
        \begin{tabular}{ccc}
            \stackunder{\includegraphics[width=0.16\textwidth]{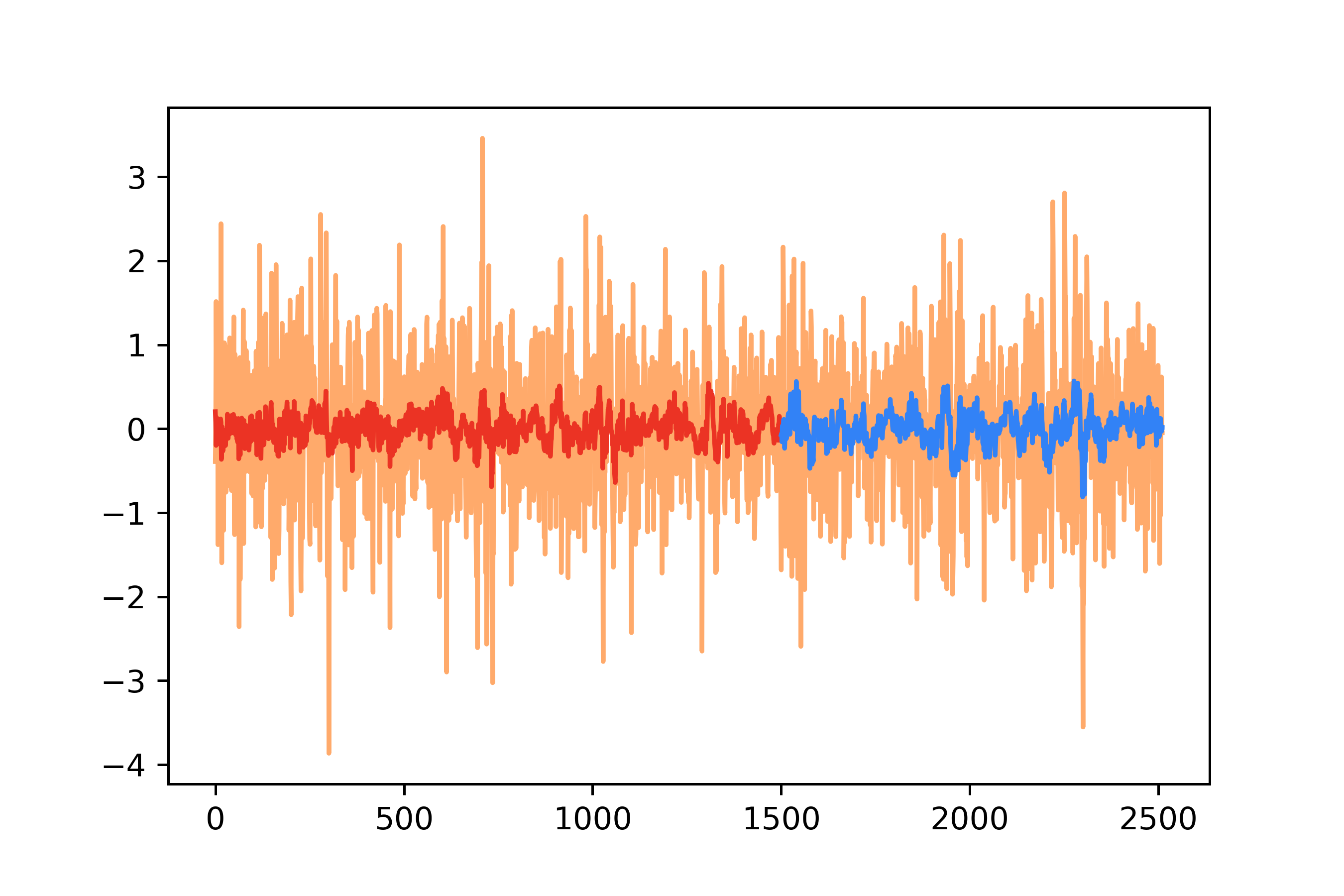}}{\scriptsize $\dot{\omega_x}$} &
            \stackunder{\includegraphics[width=0.16\textwidth]{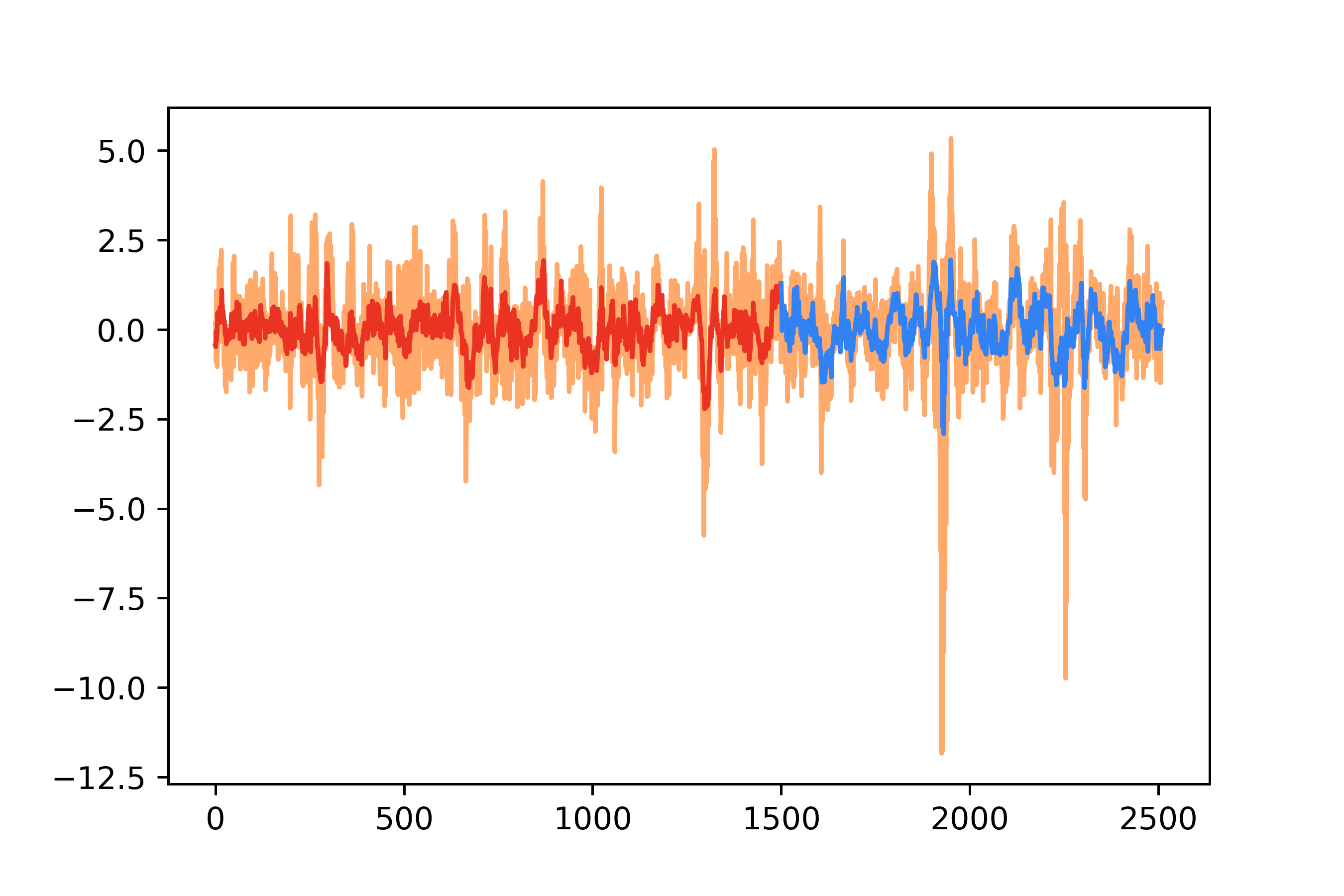}}{\scriptsize $\dot{\omega_y}$} &
            \stackunder{\includegraphics[width=0.16\textwidth]{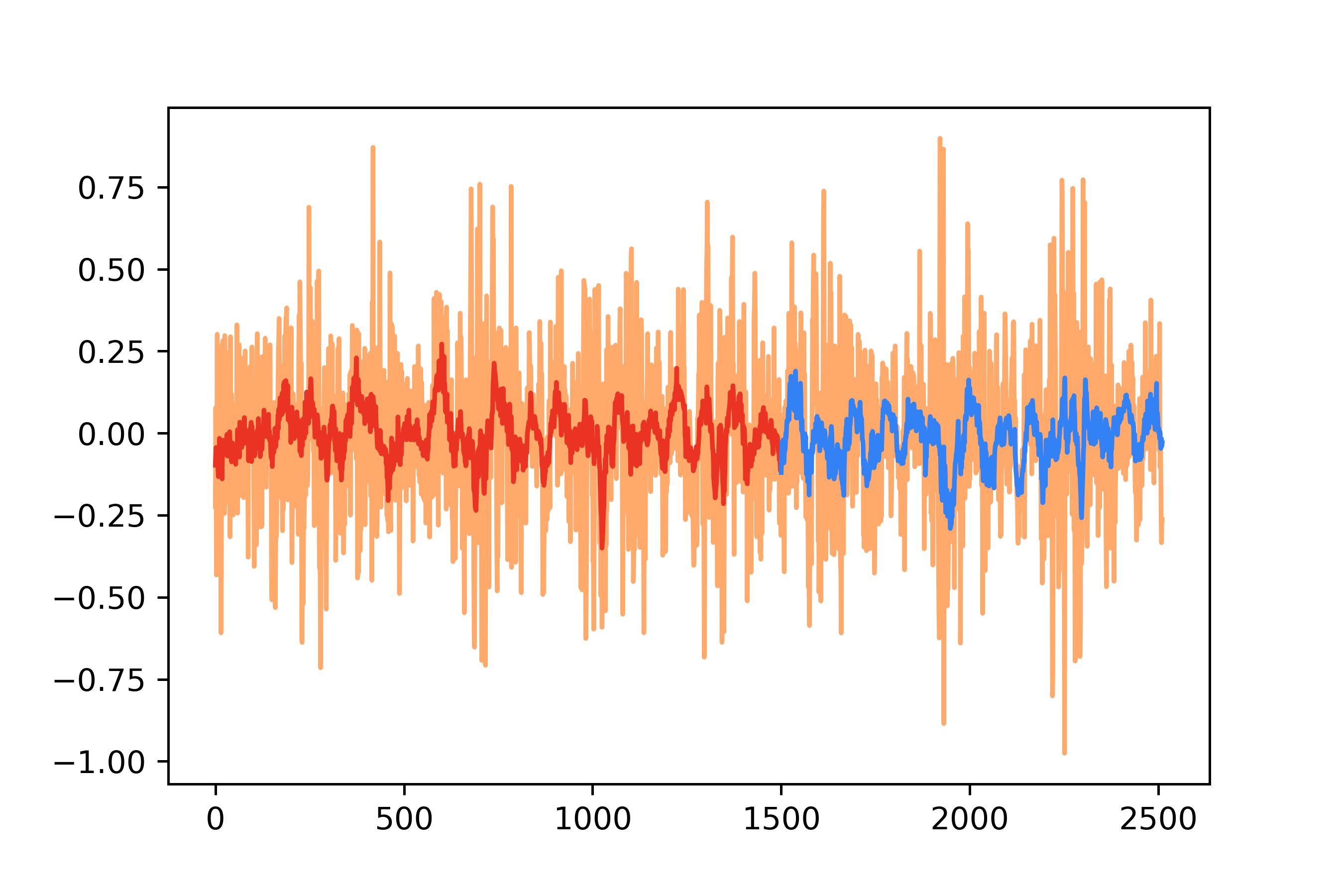}}{\scriptsize $\dot{\omega_z}$}
        \end{tabular} &
        \begin{tabular}{ccc}
            \stackunder{\includegraphics[width=0.16\textwidth]{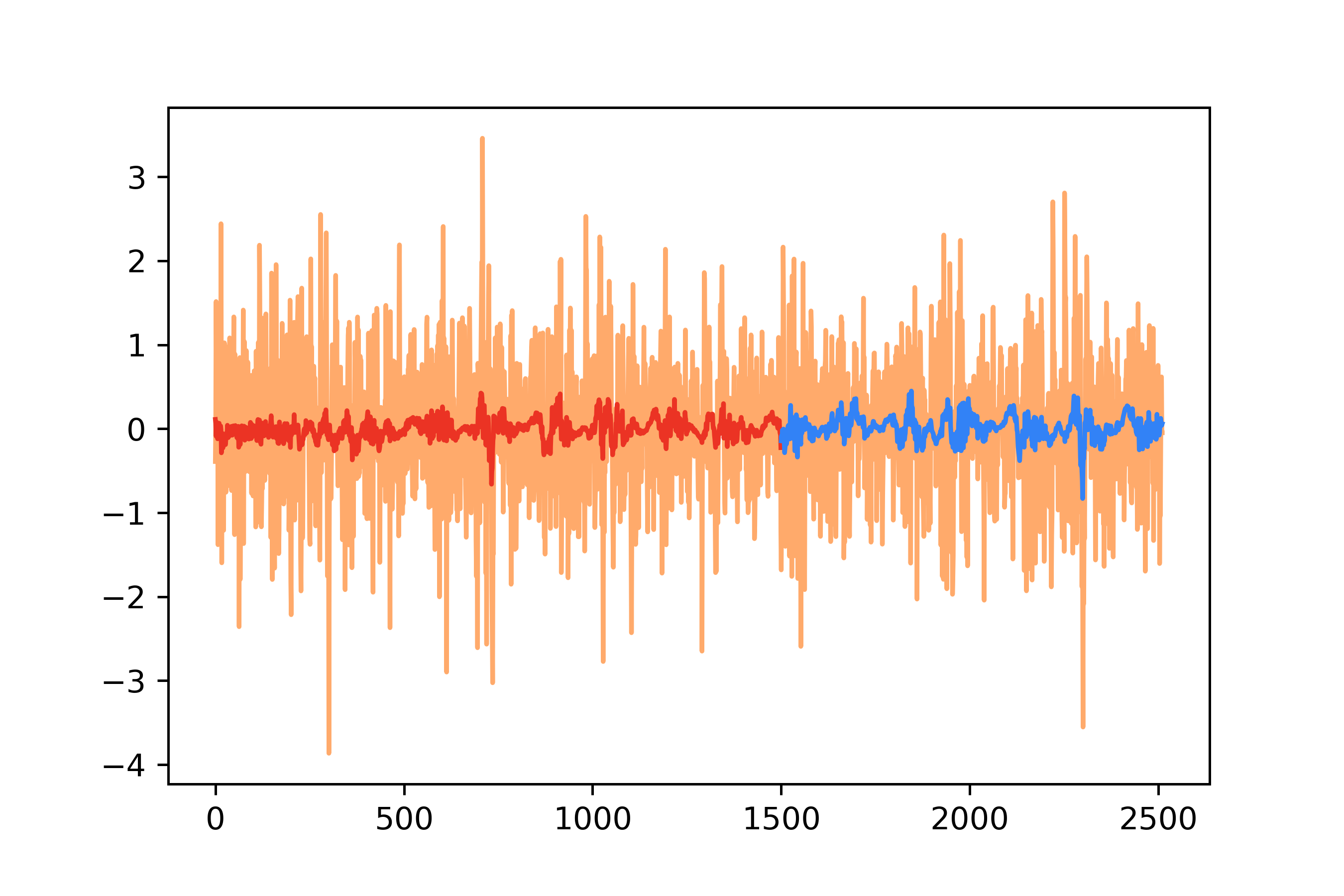}}{\scriptsize $\dot{\omega_x}$} &
            \stackunder{\includegraphics[width=0.16\textwidth]{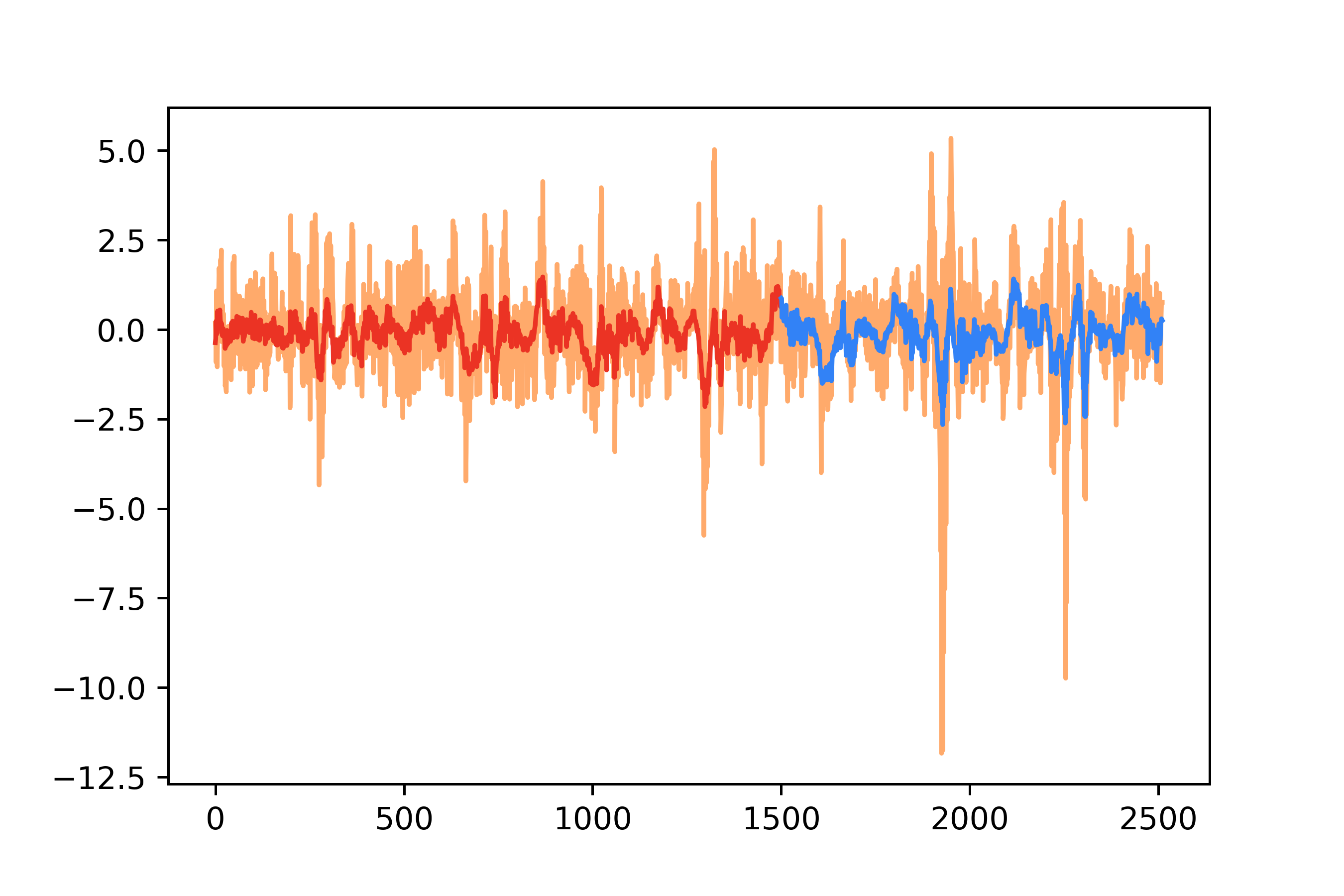}}{\scriptsize $\dot{\omega_y}$} &
            \stackunder{\includegraphics[width=0.16\textwidth]{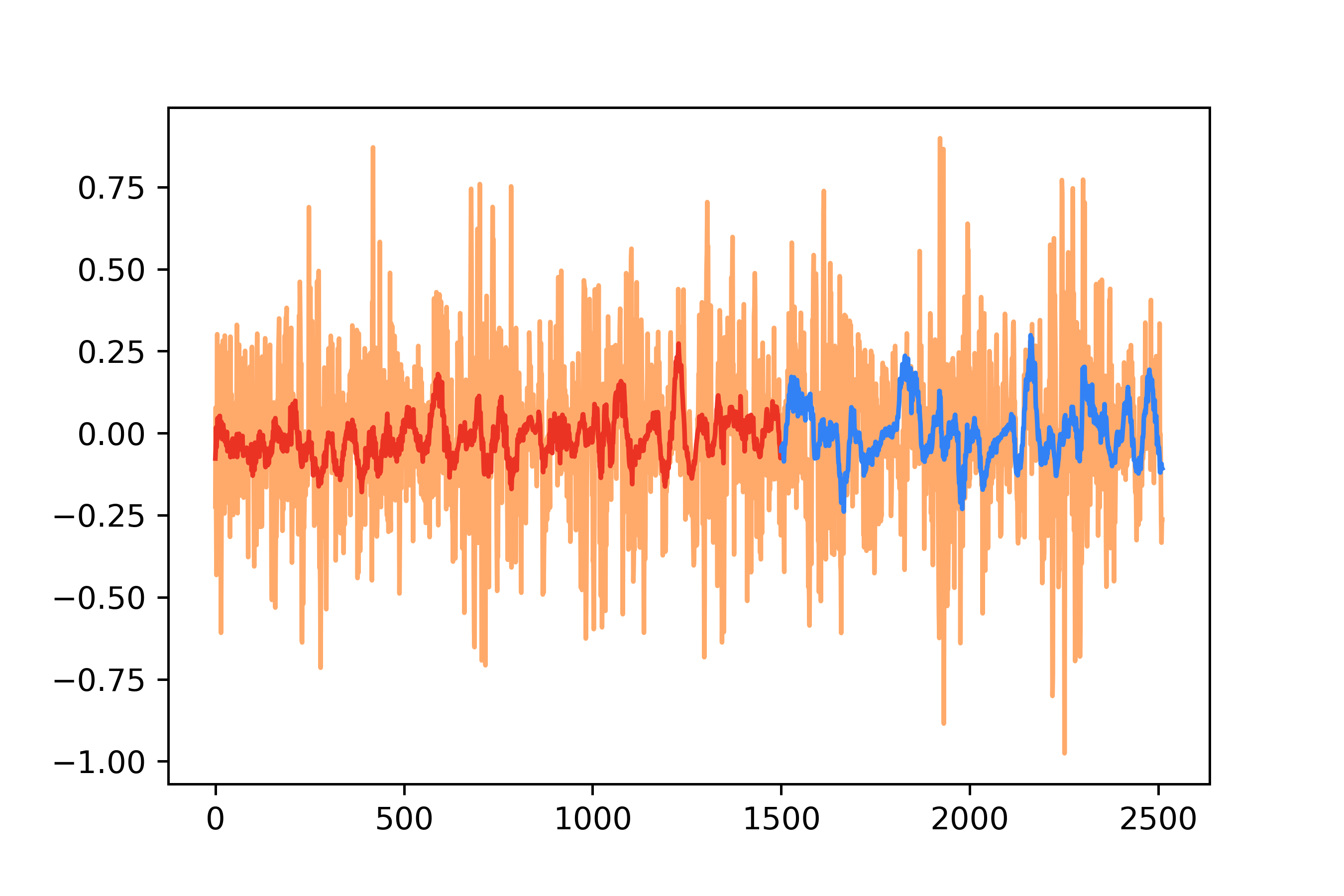}}{\scriptsize $\dot{\omega_z}$}
        \end{tabular} \\ \hline
        \small SINDy (100wind) & \small LeARN (100wind) \\ \hline
        \begin{tabular}{ccc}
            \stackunder{\includegraphics[width=0.16\textwidth]{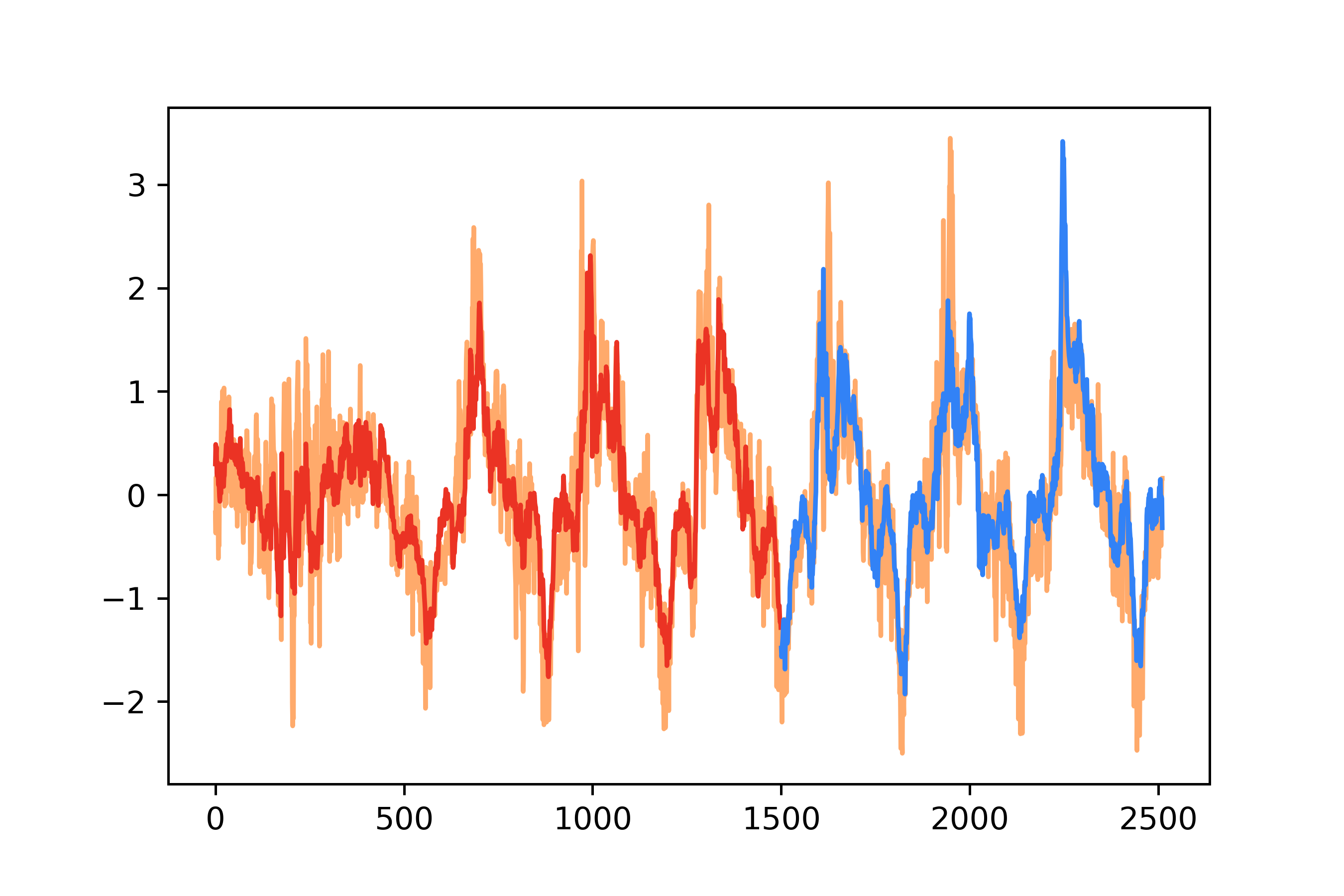}}{\scriptsize $\dot{v_x}$} &
            \stackunder{\includegraphics[width=0.16\textwidth]{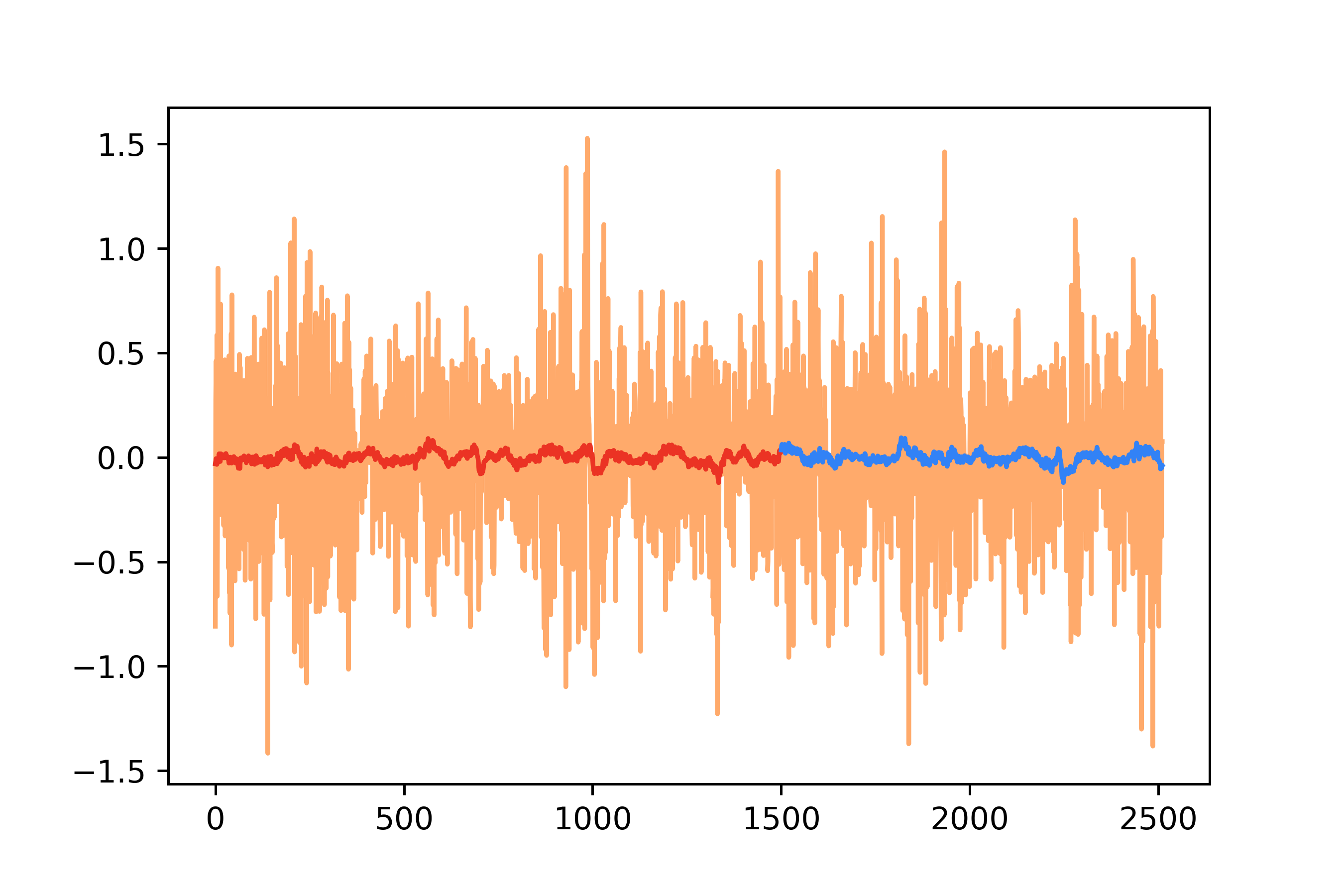}}{\scriptsize $\dot{v_y}$} &
            \stackunder{\includegraphics[width=0.16\textwidth]{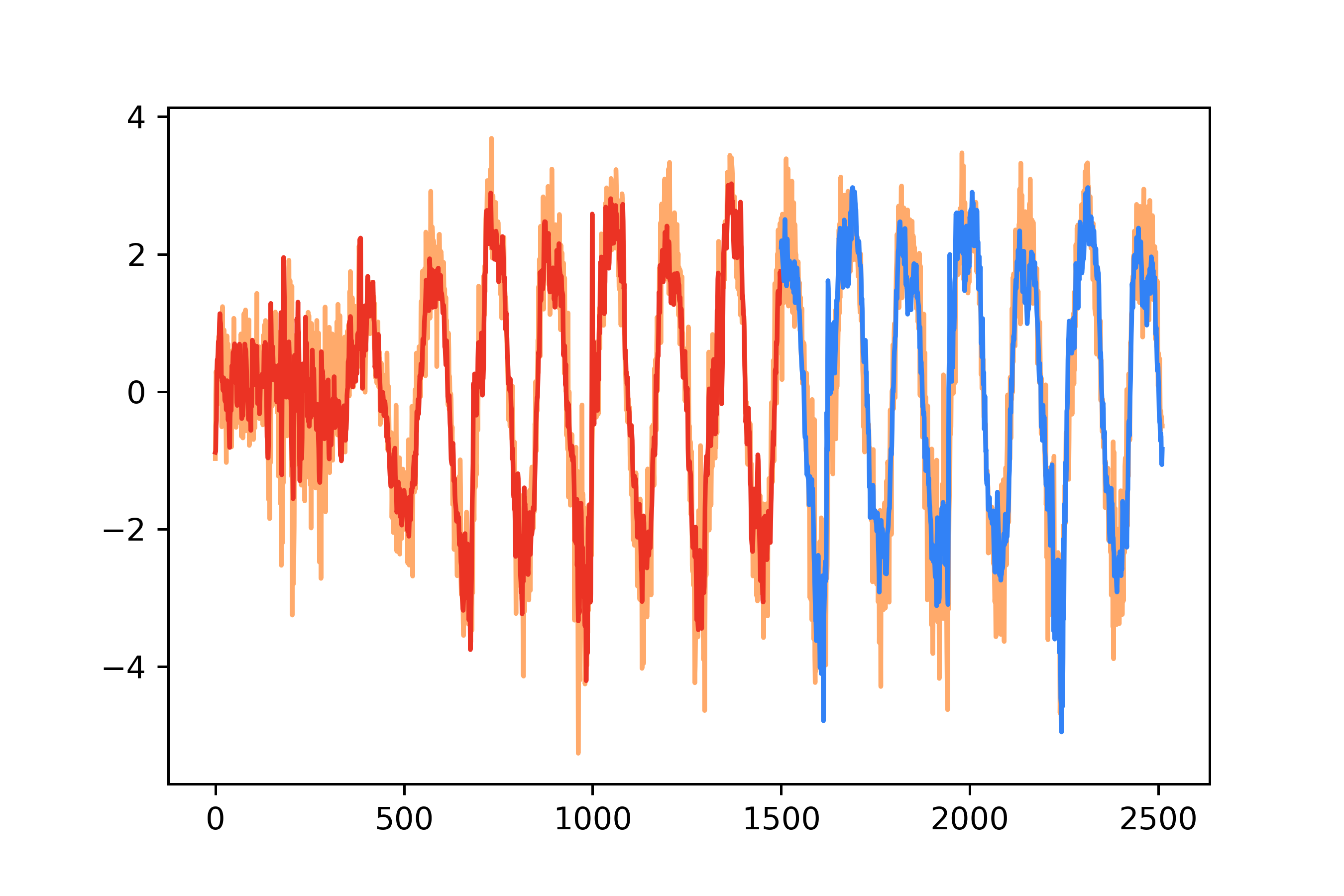}}{\scriptsize $\dot{v_z}$}
        \end{tabular} &
        \begin{tabular}{ccc}
            \stackunder{\includegraphics[width=0.16\textwidth]{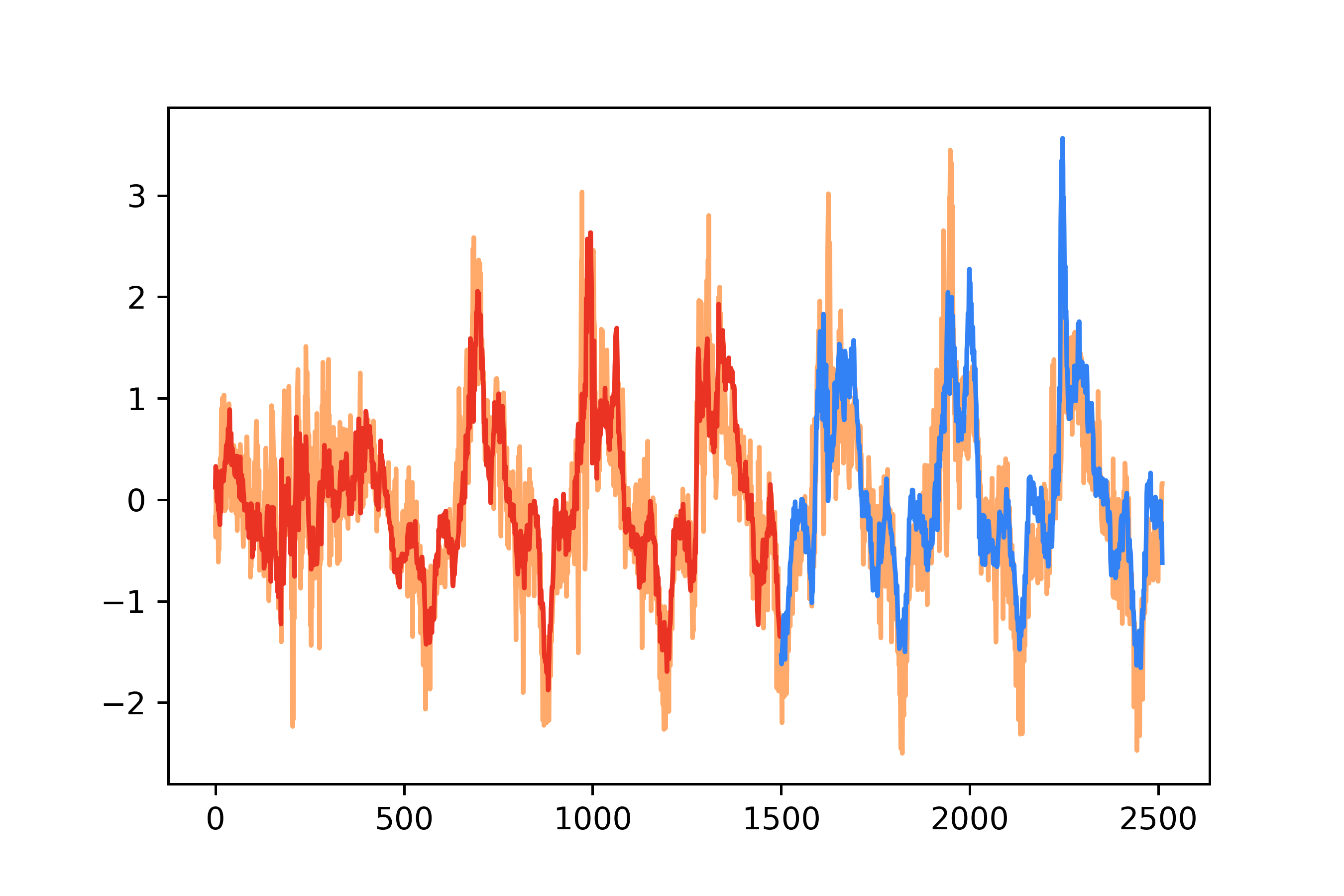}}{\scriptsize $\dot{v_x}$} &
            \stackunder{\includegraphics[width=0.16\textwidth]{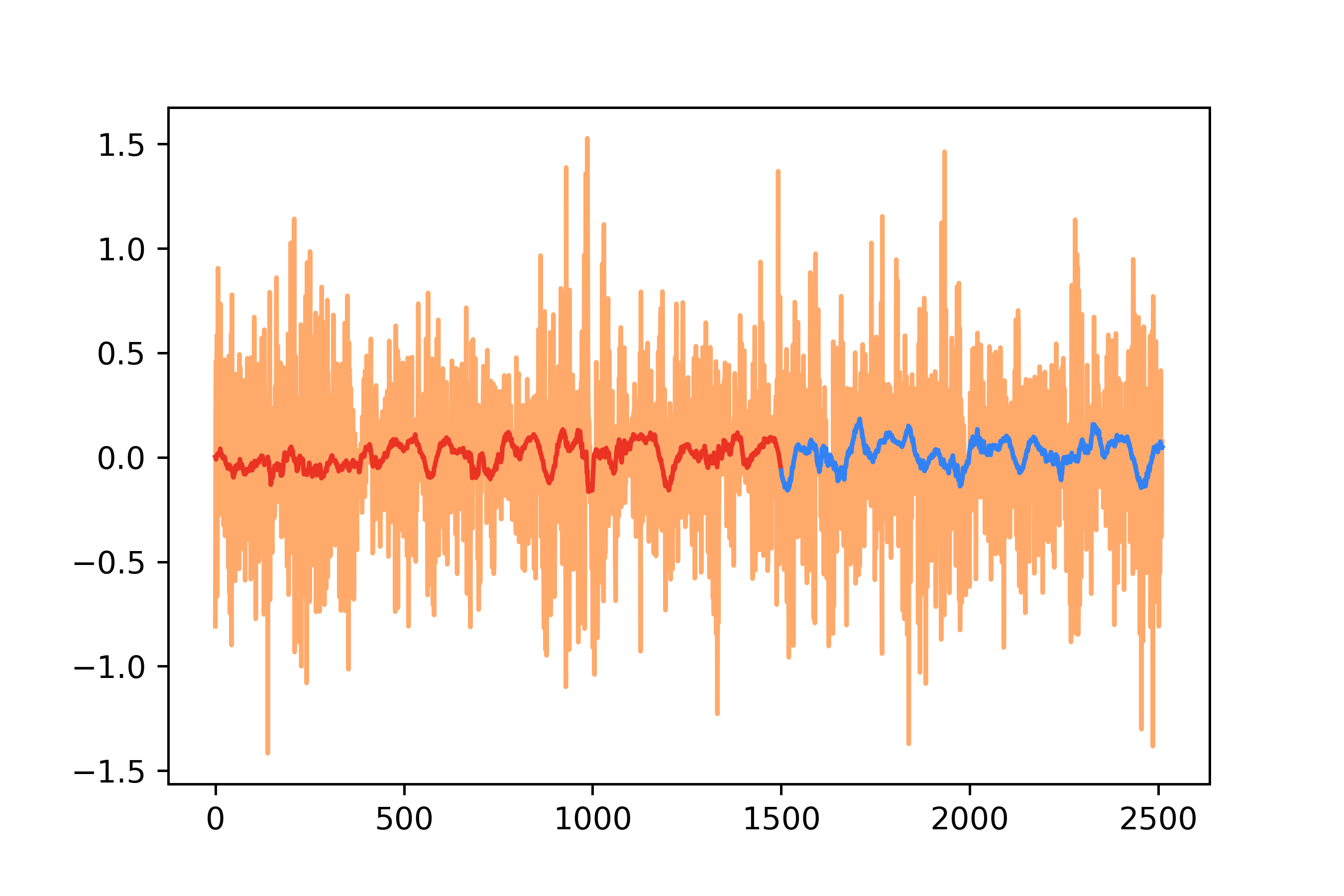}}{\scriptsize $\dot{v_y}$} &
            \stackunder{\includegraphics[width=0.16\textwidth]{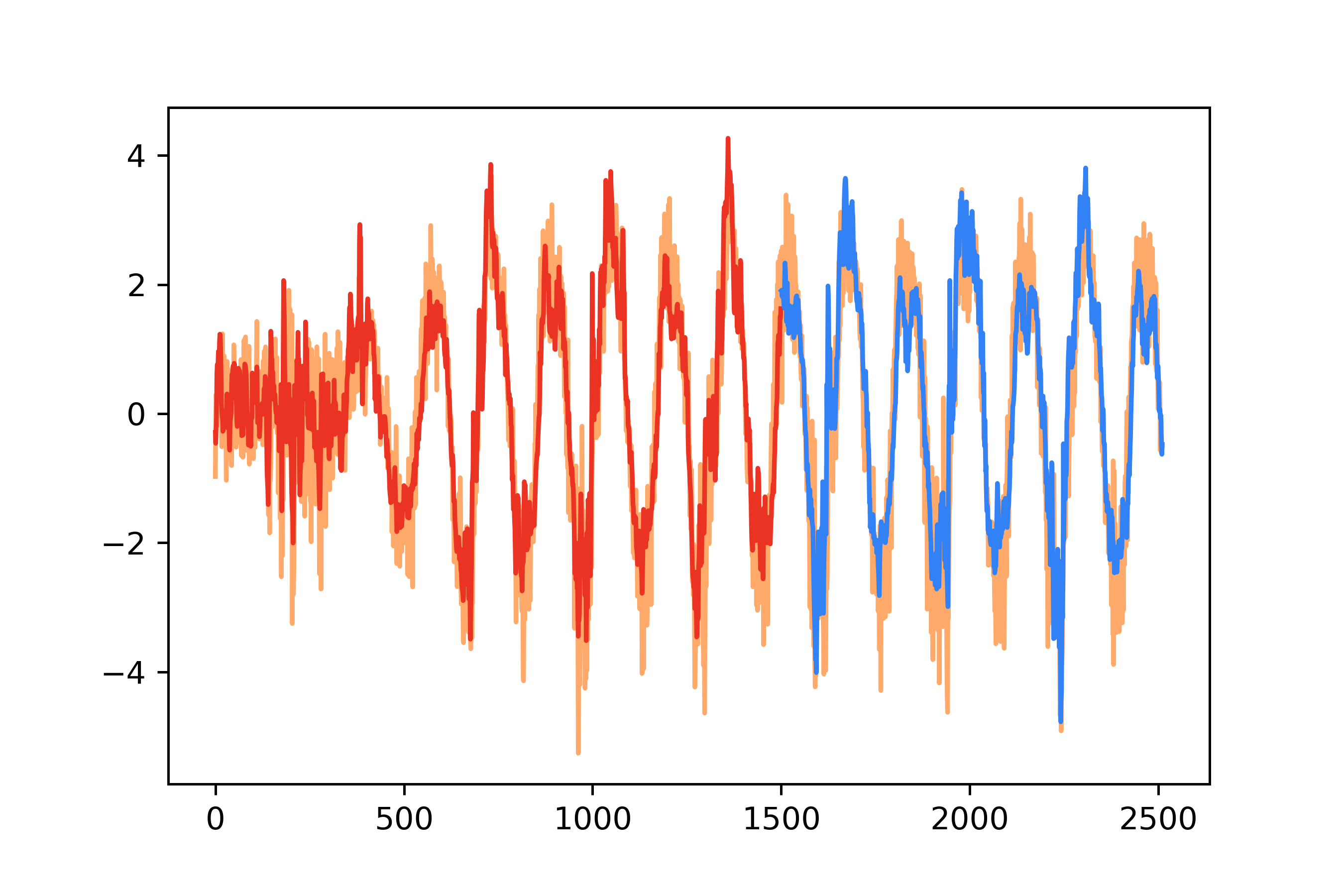}}{\scriptsize $\dot{v_z}$}
        \end{tabular} \\ \hline
        \begin{tabular}{ccc}
            \stackunder{\includegraphics[width=0.16\textwidth]{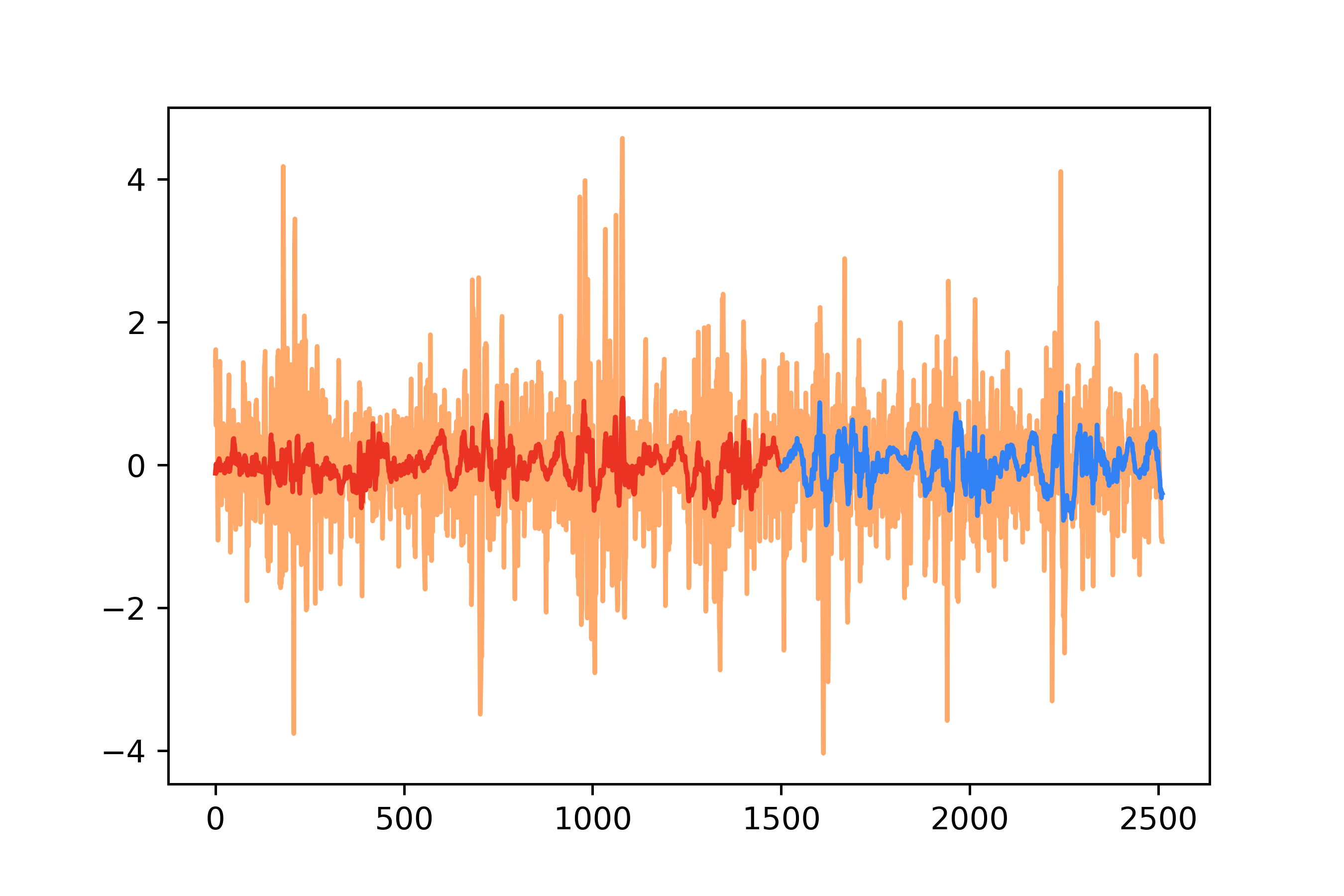}}{\scriptsize $\dot{\omega_x}$} &
            \stackunder{\includegraphics[width=0.16\textwidth]{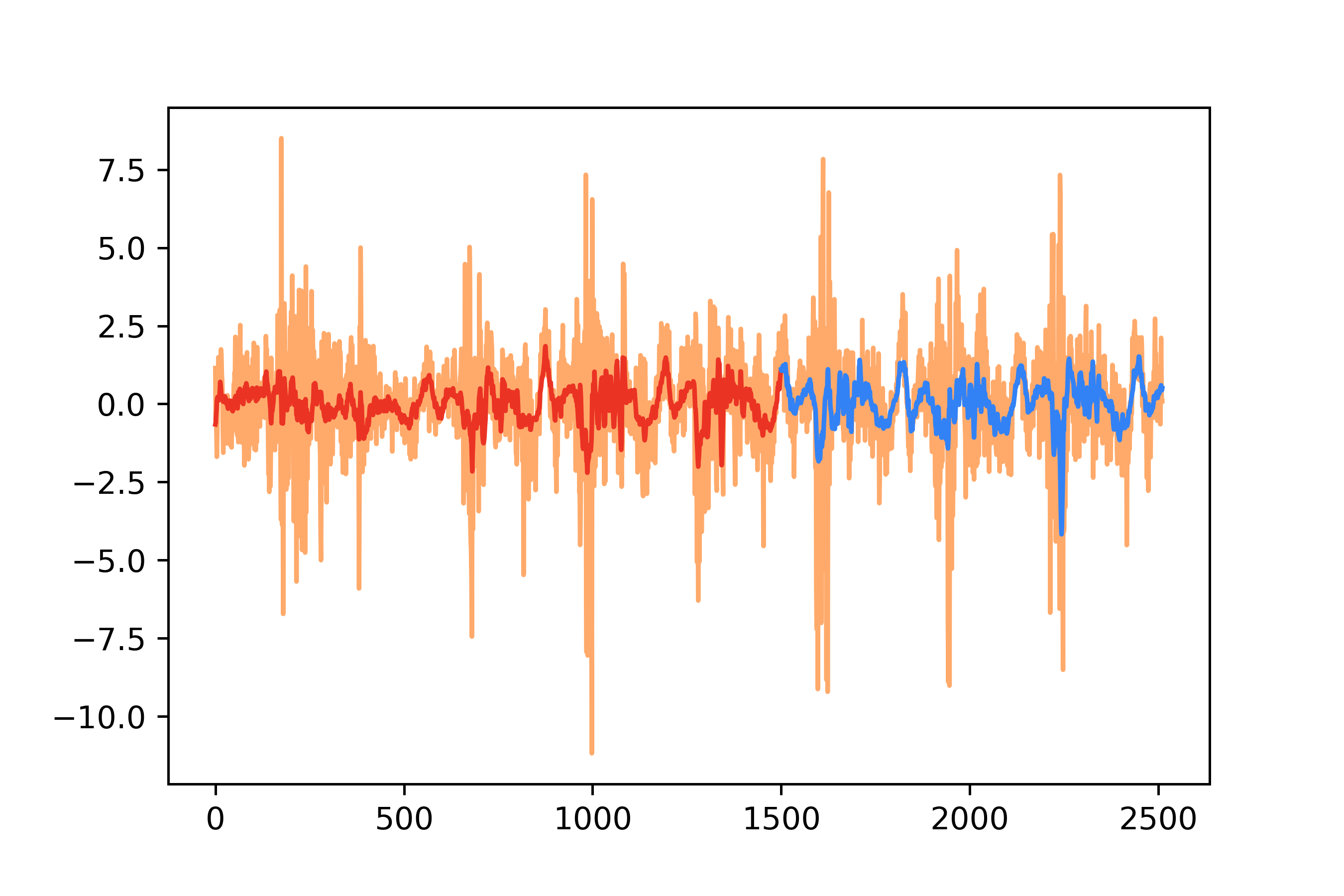}}{\scriptsize $\dot{\omega_y}$} &
            \stackunder{\includegraphics[width=0.16\textwidth]{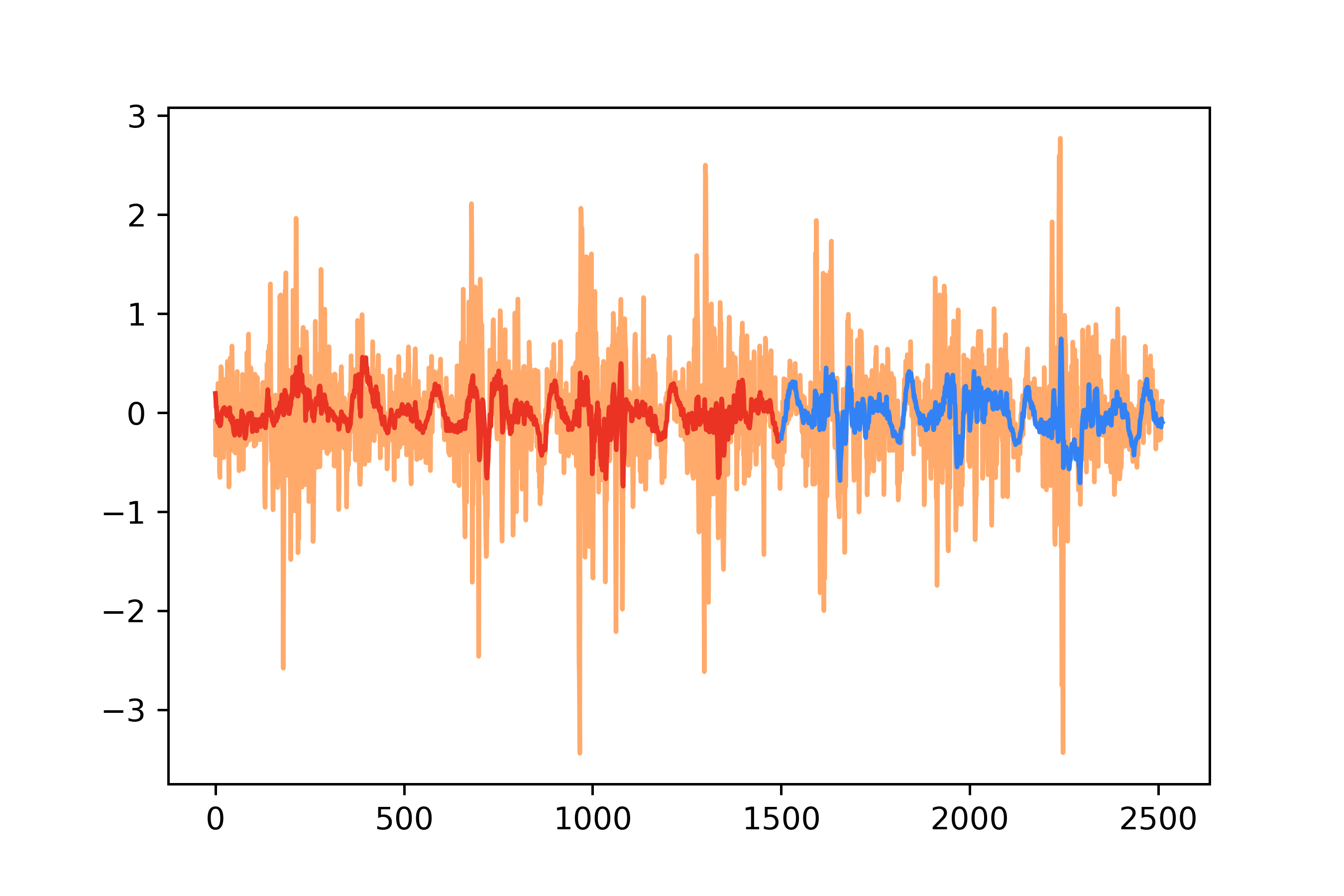}}{\scriptsize $\dot{\omega_z}$}
        \end{tabular} &
        \begin{tabular}{ccc}
            \stackunder{\includegraphics[width=0.16\textwidth]{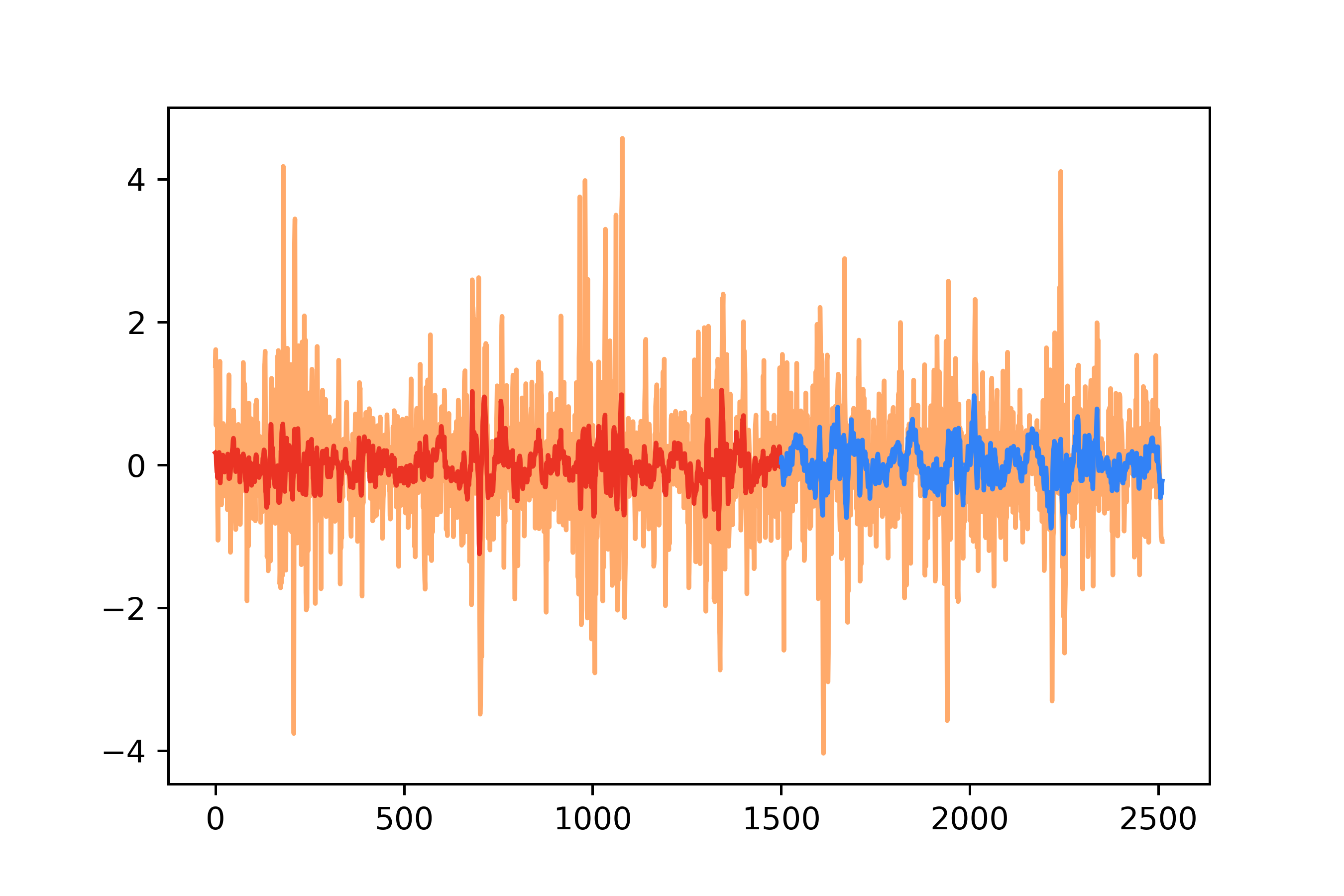}}{\scriptsize $\dot{\omega_x}$} &
            \stackunder{\includegraphics[width=0.16\textwidth]{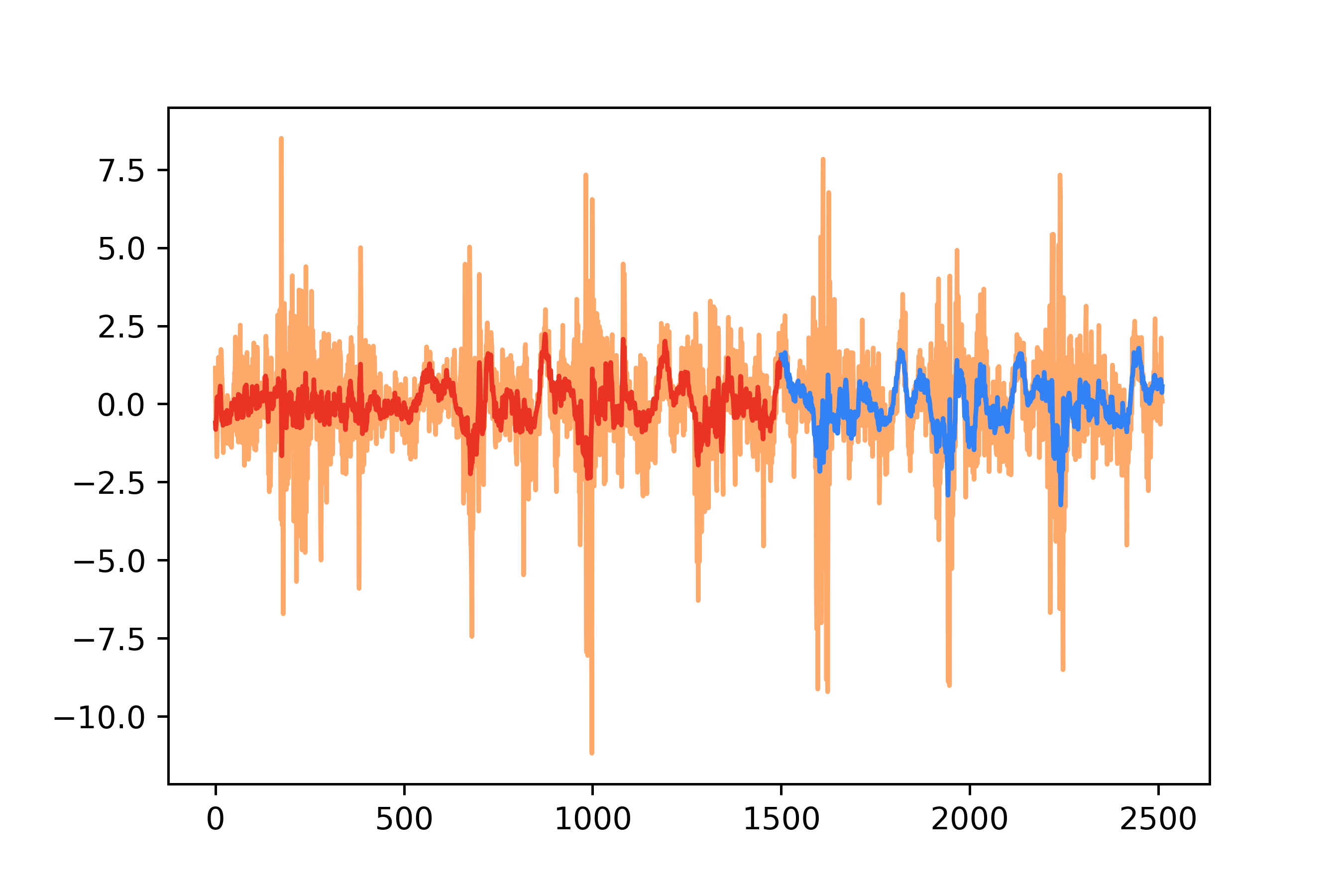}}{\scriptsize $\dot{\omega_y}$} &
            \stackunder{\includegraphics[width=0.16\textwidth]{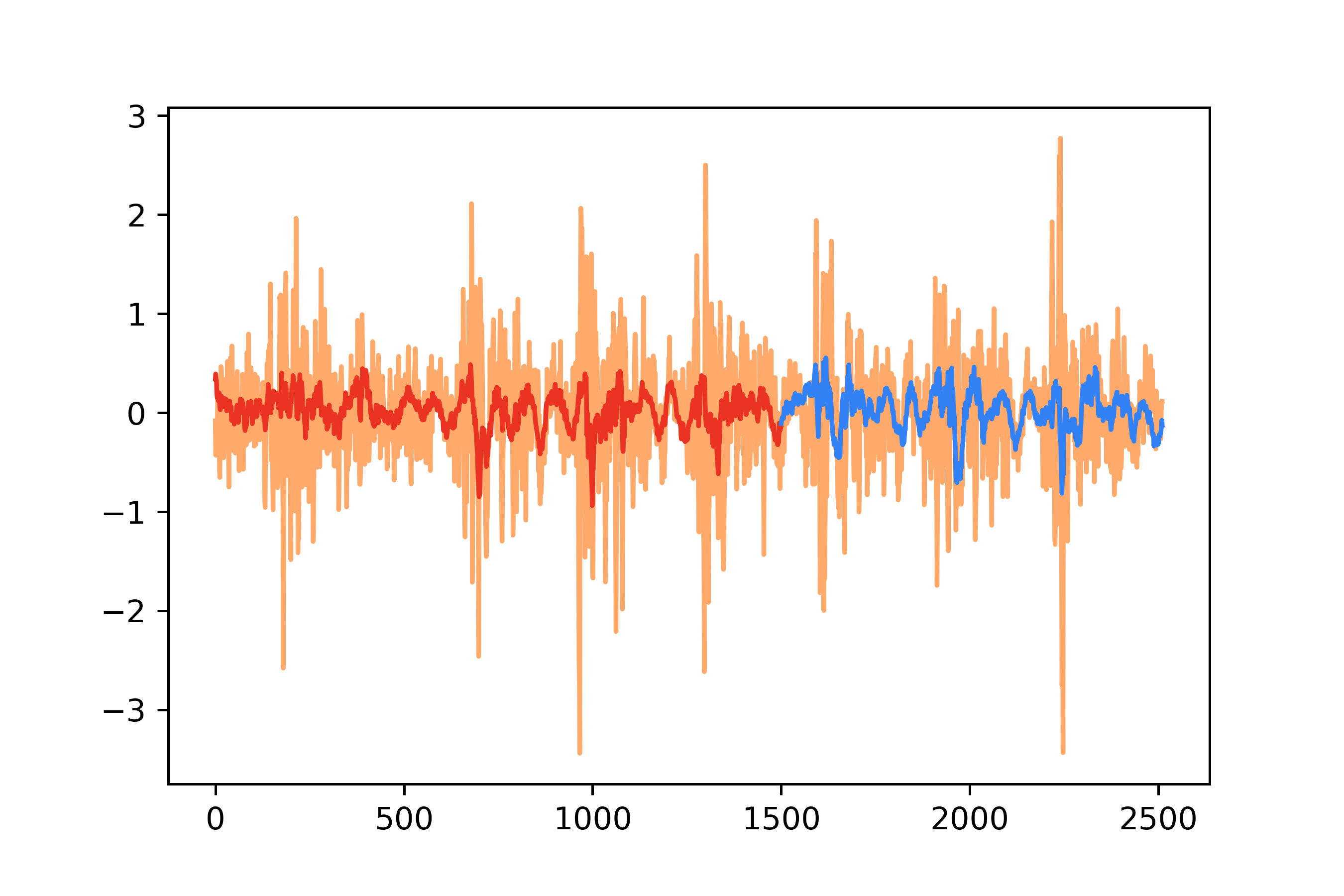}}{\scriptsize $\dot{\omega_z}$}
        \end{tabular} \\ \hline
        \small SINDy (70psin20) & \small LeARN (70psin20) \\ \hline
        \begin{tabular}{ccc}
            \stackunder{\includegraphics[width=0.16\textwidth]{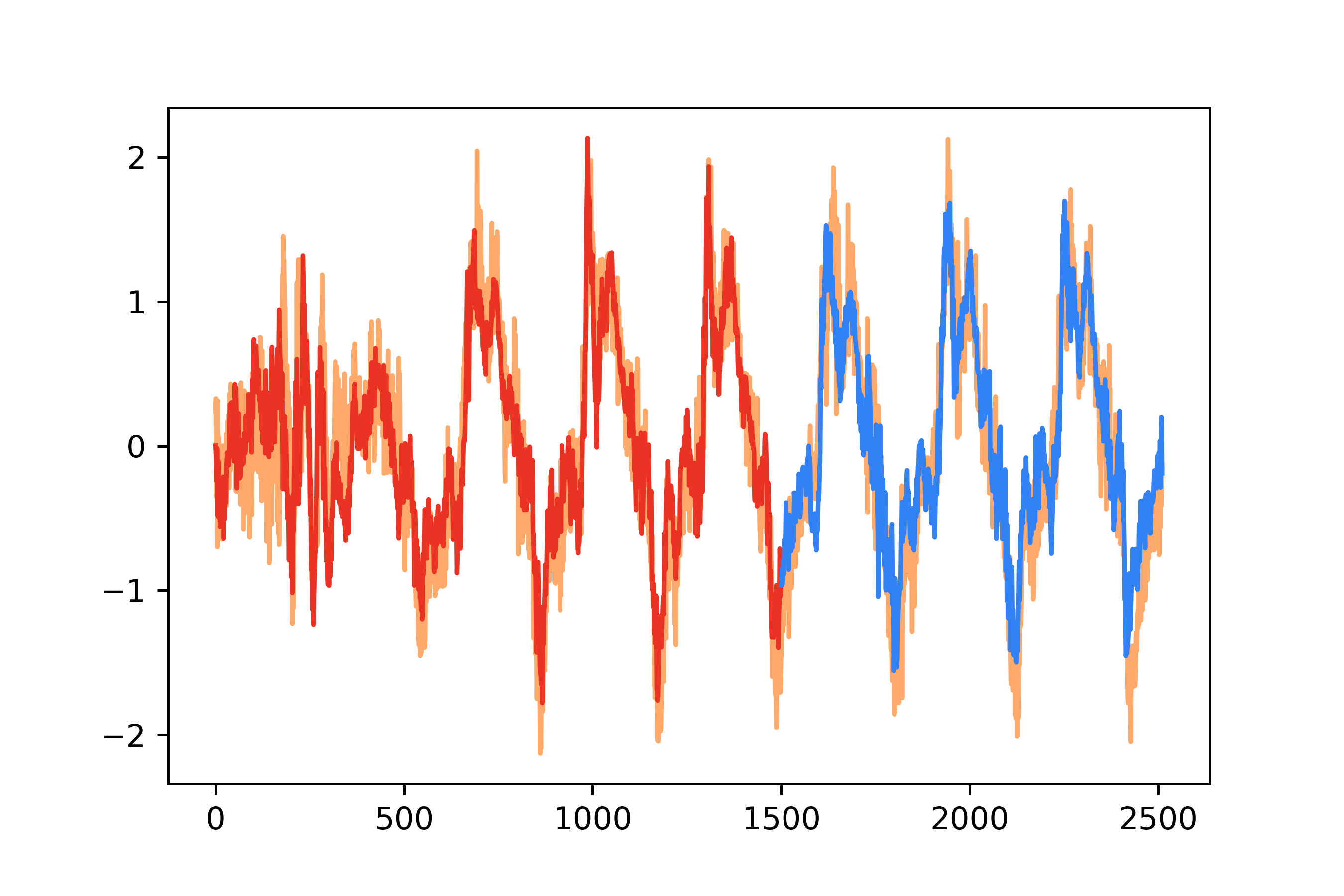}}{\scriptsize $\dot{v_x}$} &
            \stackunder{\includegraphics[width=0.16\textwidth]{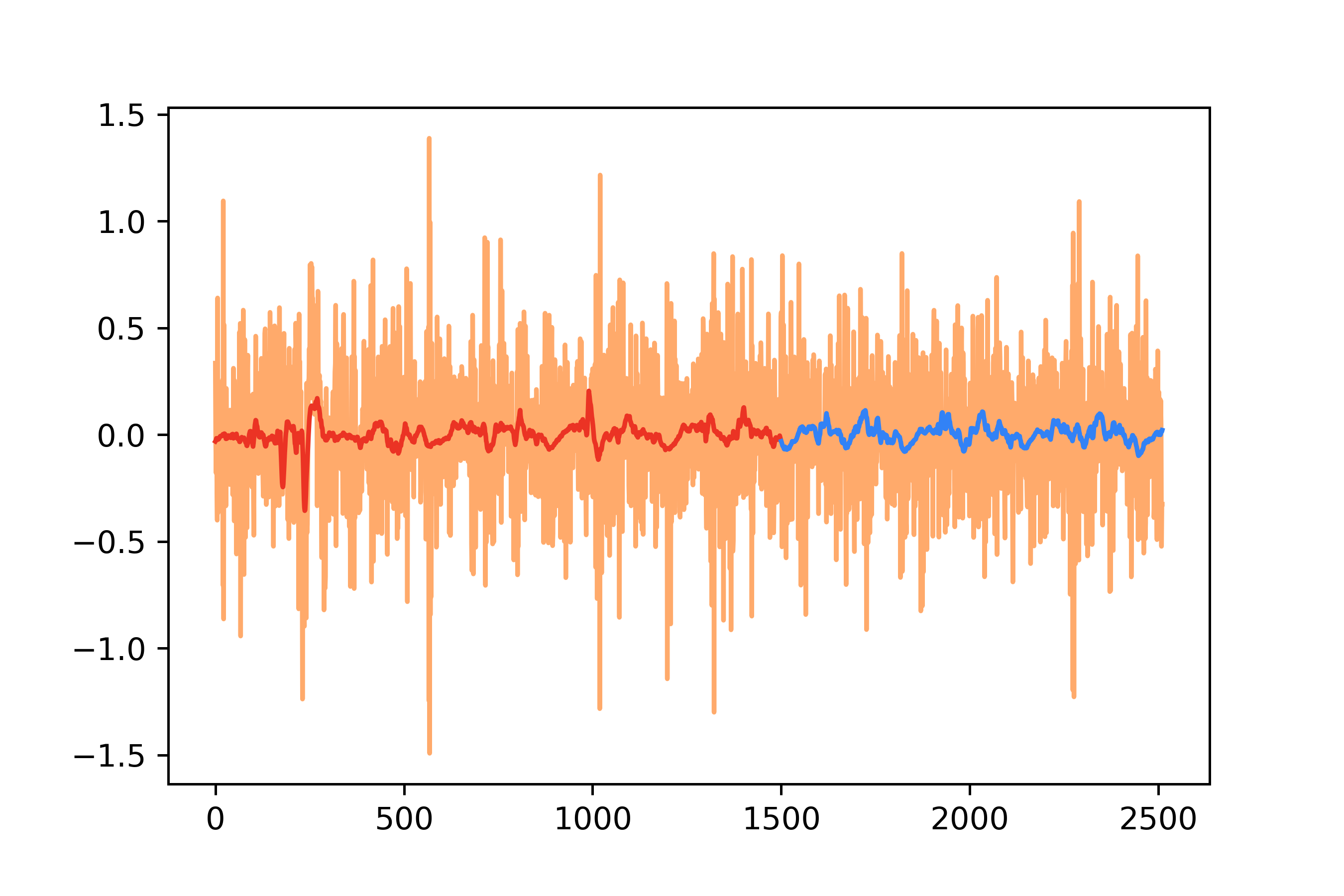}}{\scriptsize $\dot{v_y}$} &
            \stackunder{\includegraphics[width=0.16\textwidth]{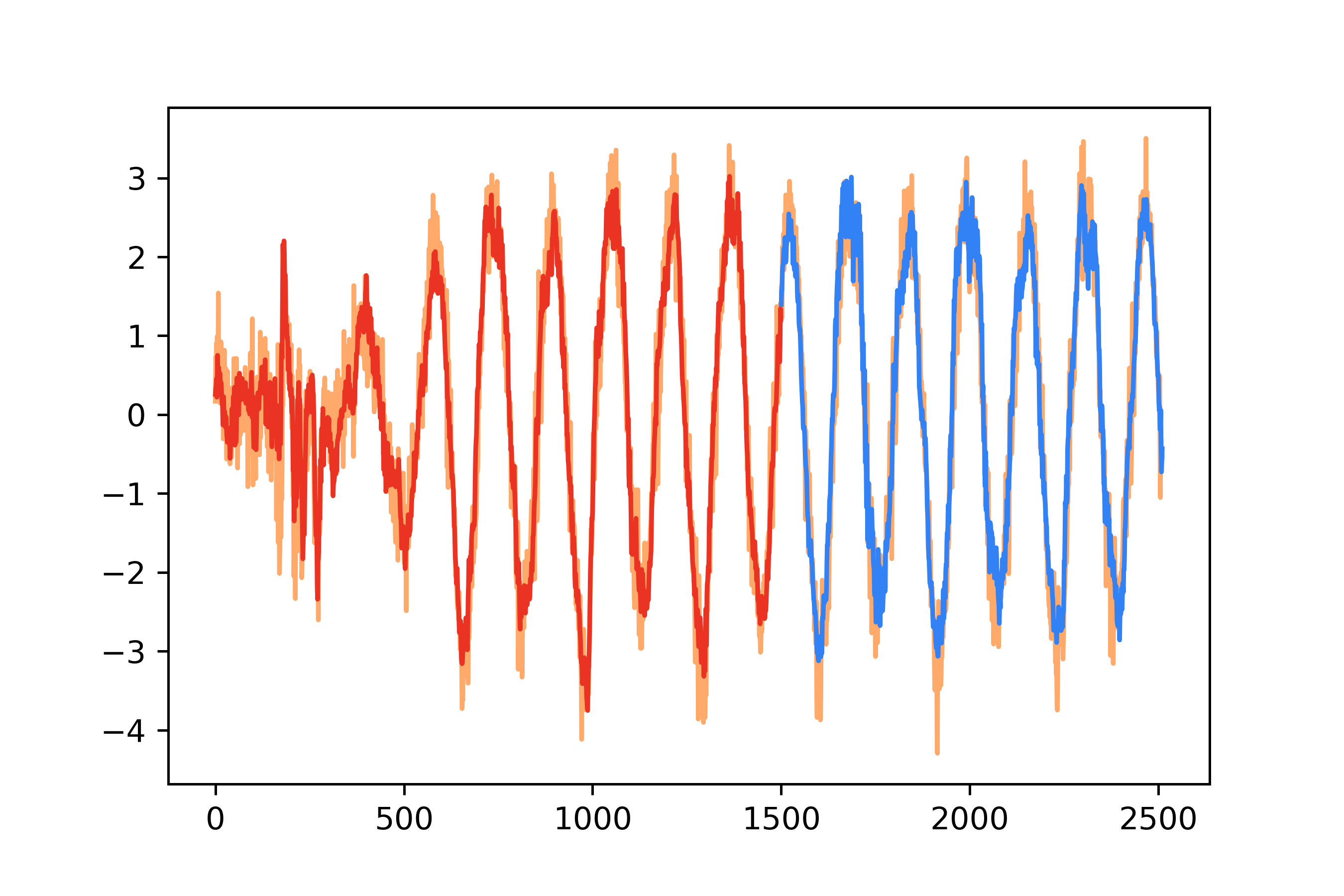}}{\scriptsize $\dot{v_z}$}
        \end{tabular} &
        \begin{tabular}{ccc}
            \stackunder{\includegraphics[width=0.16\textwidth]{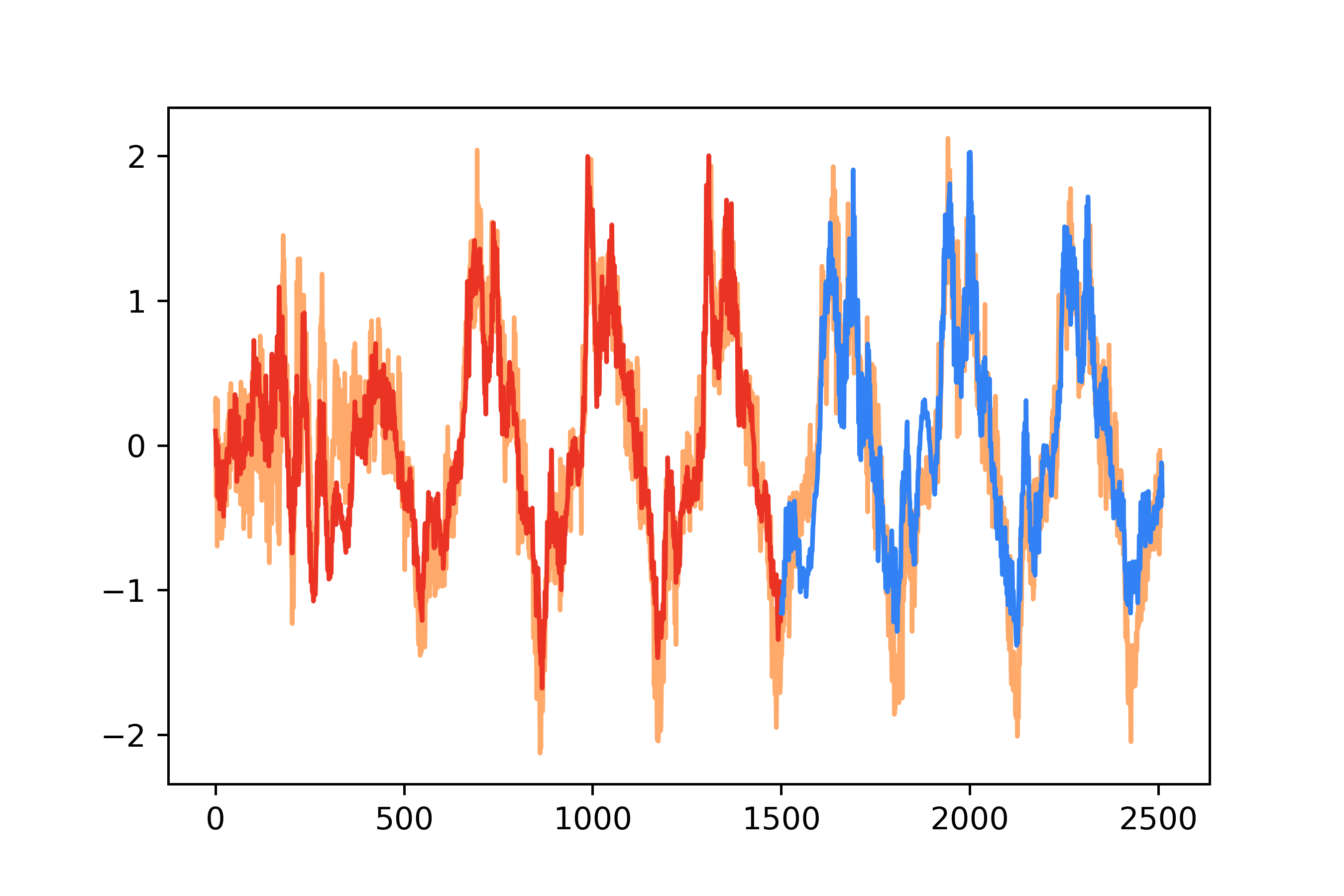}}{\scriptsize $\dot{v_x}$} &
            \stackunder{\includegraphics[width=0.16\textwidth]{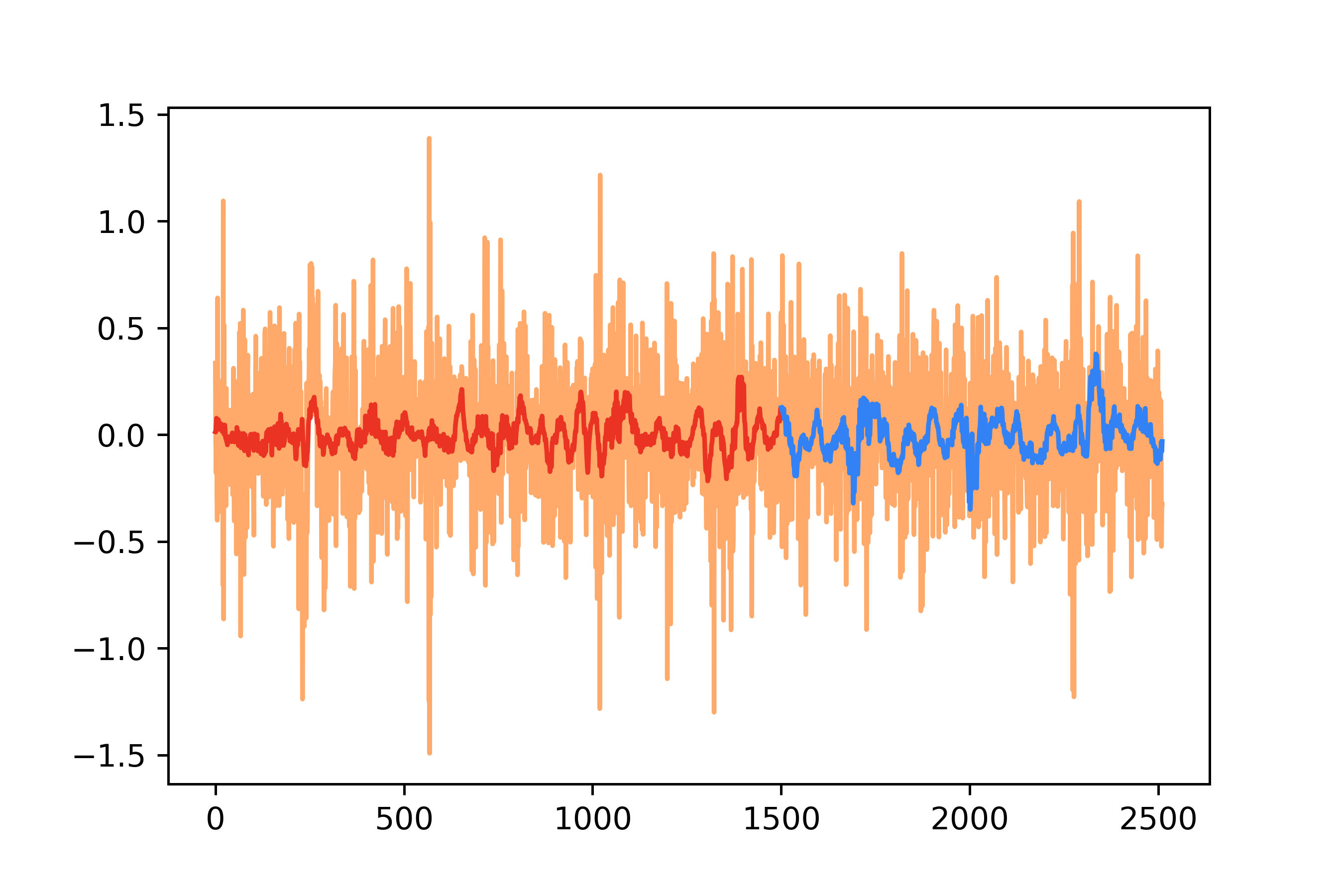}}{\scriptsize $\dot{v_y}$} &
            \stackunder{\includegraphics[width=0.16\textwidth]{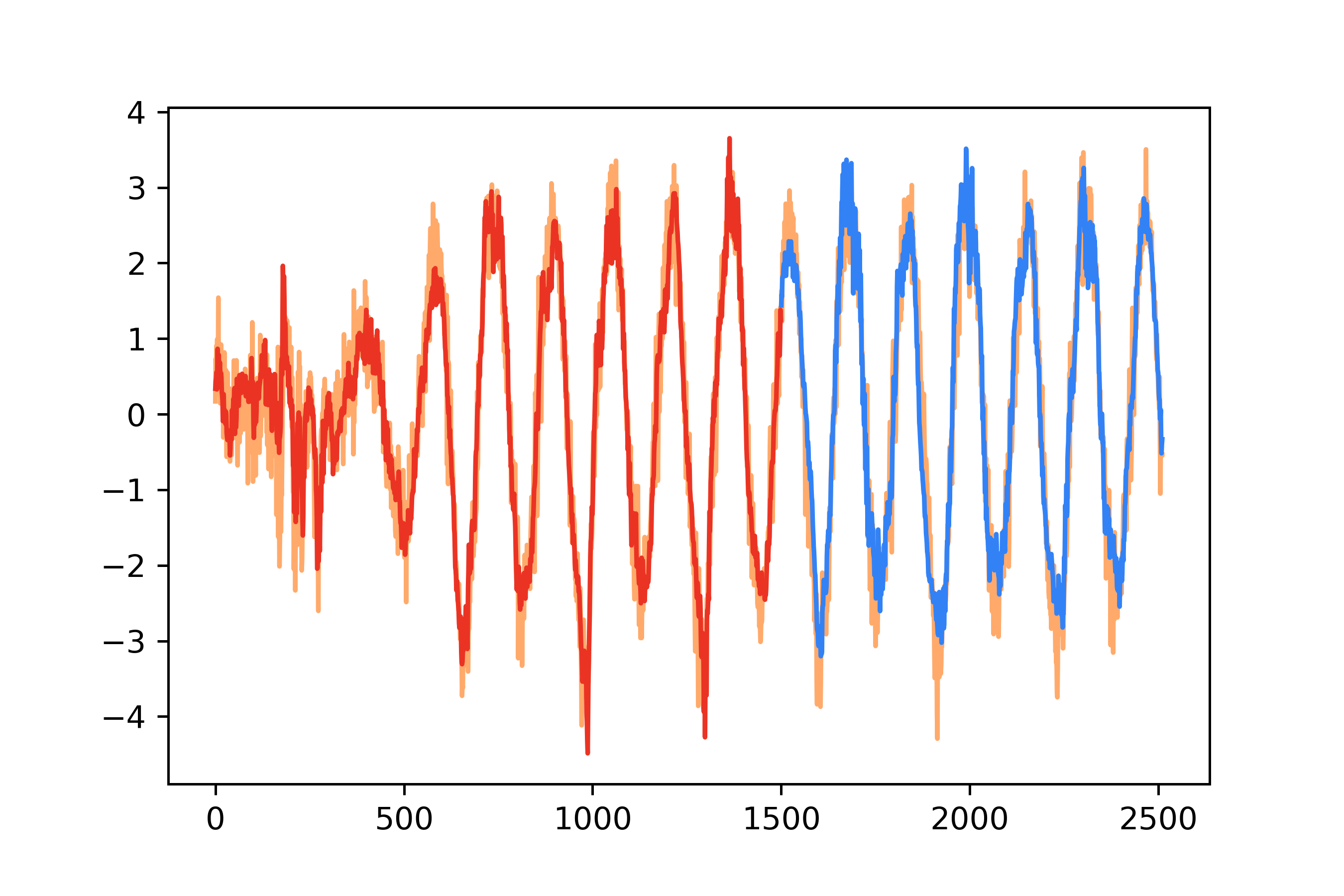}}{\scriptsize $\dot{v_z}$}
        \end{tabular} \\ \hline
        \begin{tabular}{ccc}
            \stackunder{\includegraphics[width=0.16\textwidth]{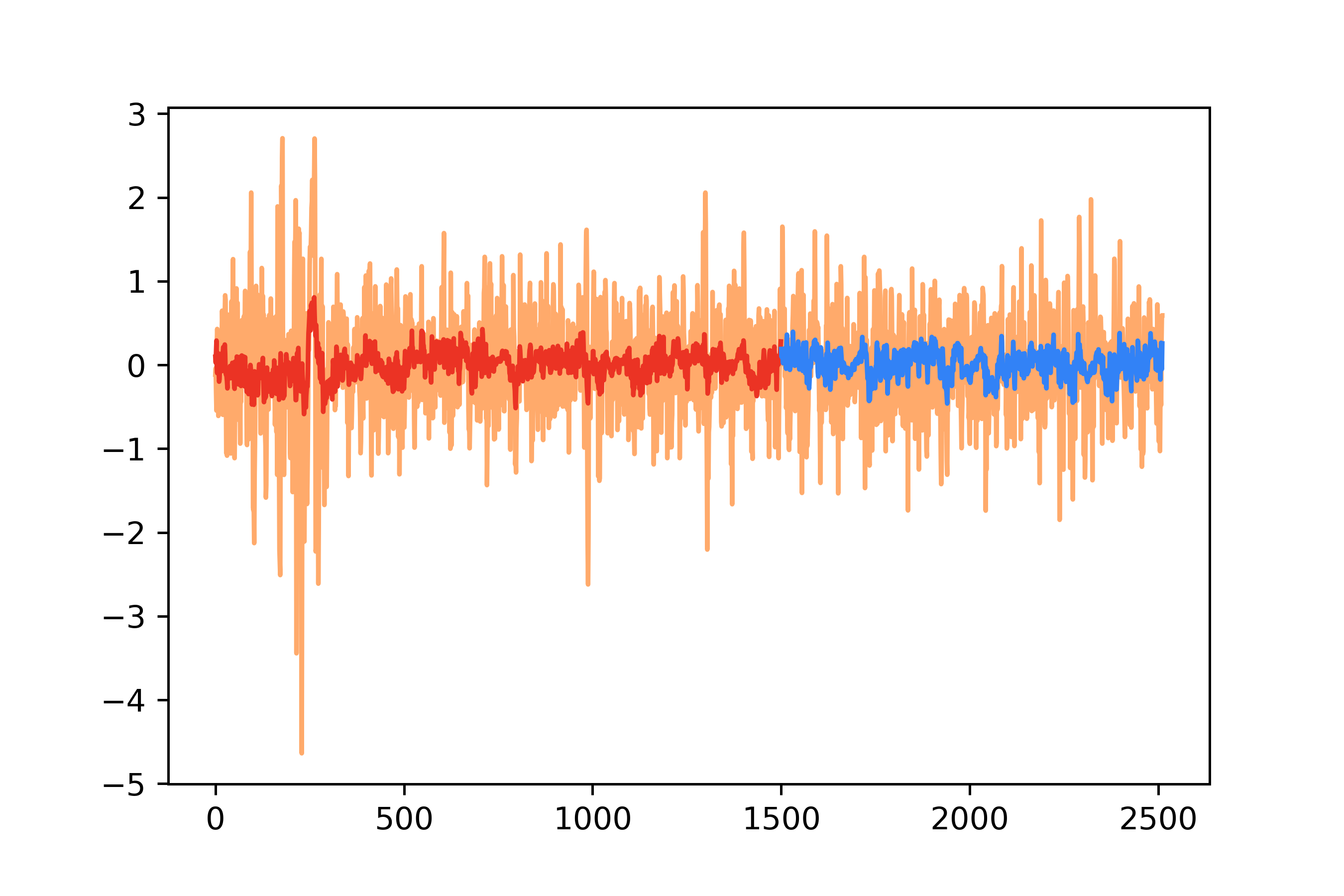}}{\scriptsize $\dot{\omega_x}$} &
            \stackunder{\includegraphics[width=0.16\textwidth]{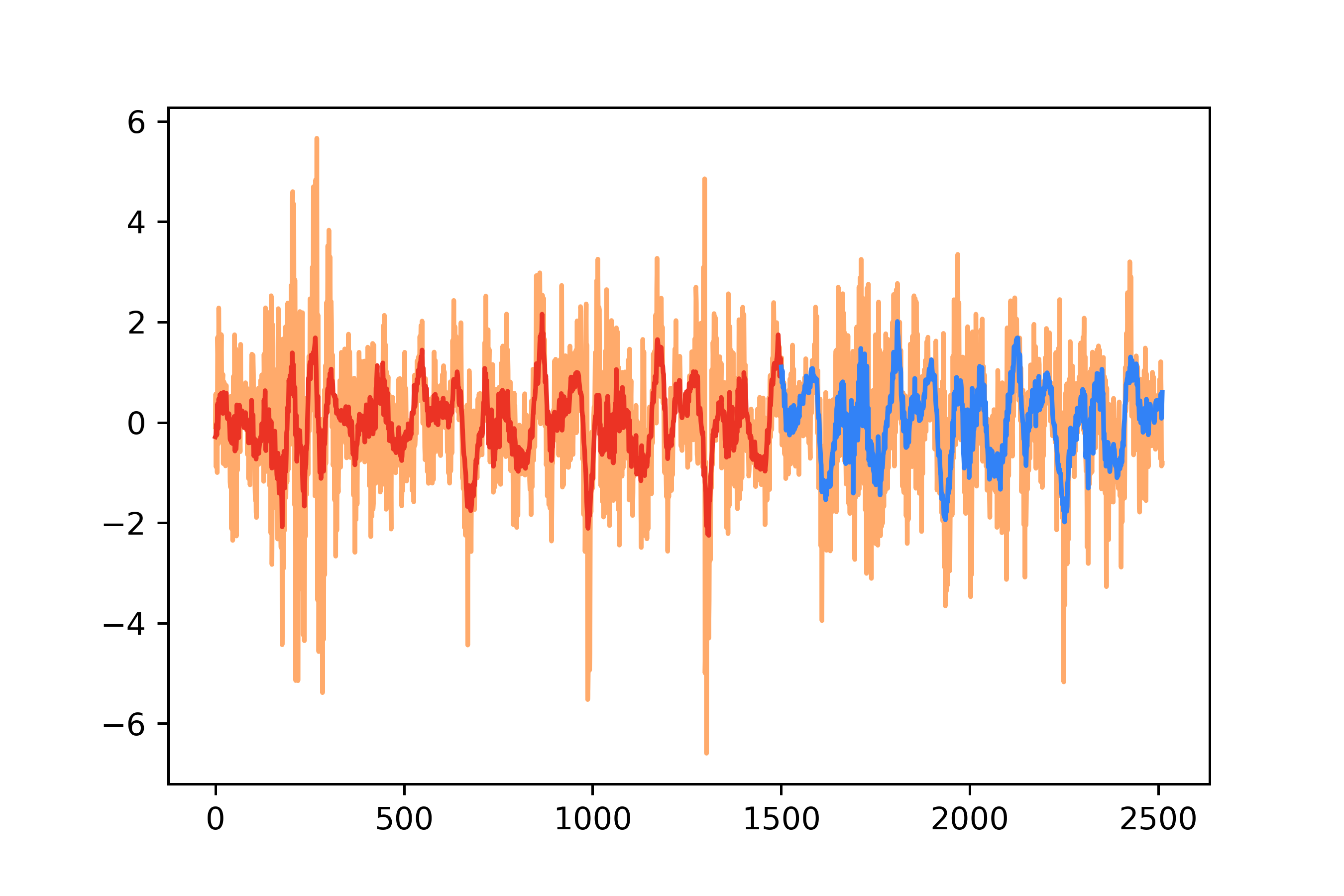}}{\scriptsize $\dot{\omega_y}$} &
            \stackunder{\includegraphics[width=0.16\textwidth]{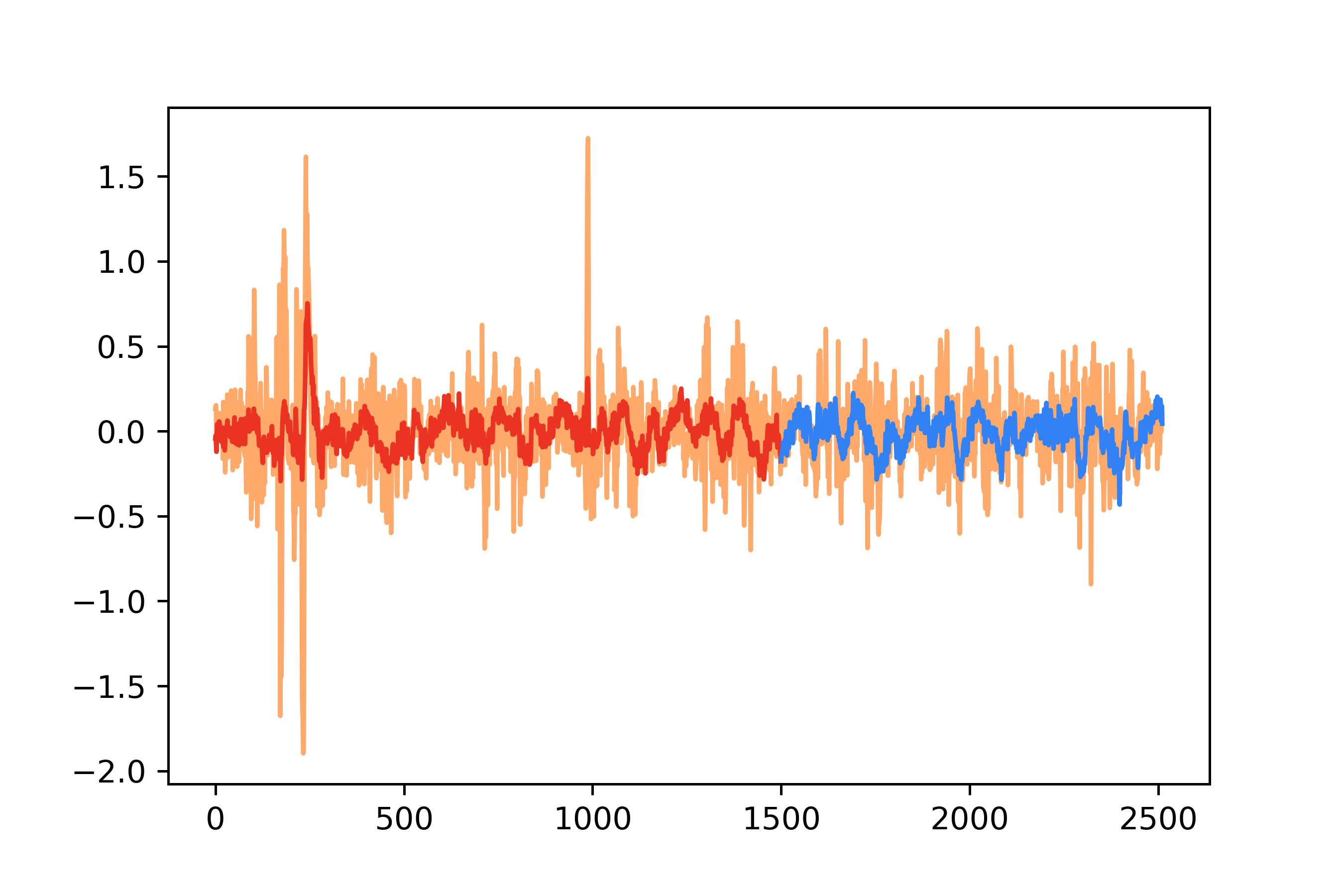}}{\scriptsize $\dot{\omega_z}$}
        \end{tabular} &
        \begin{tabular}{ccc}
            \stackunder{\includegraphics[width=0.16\textwidth]{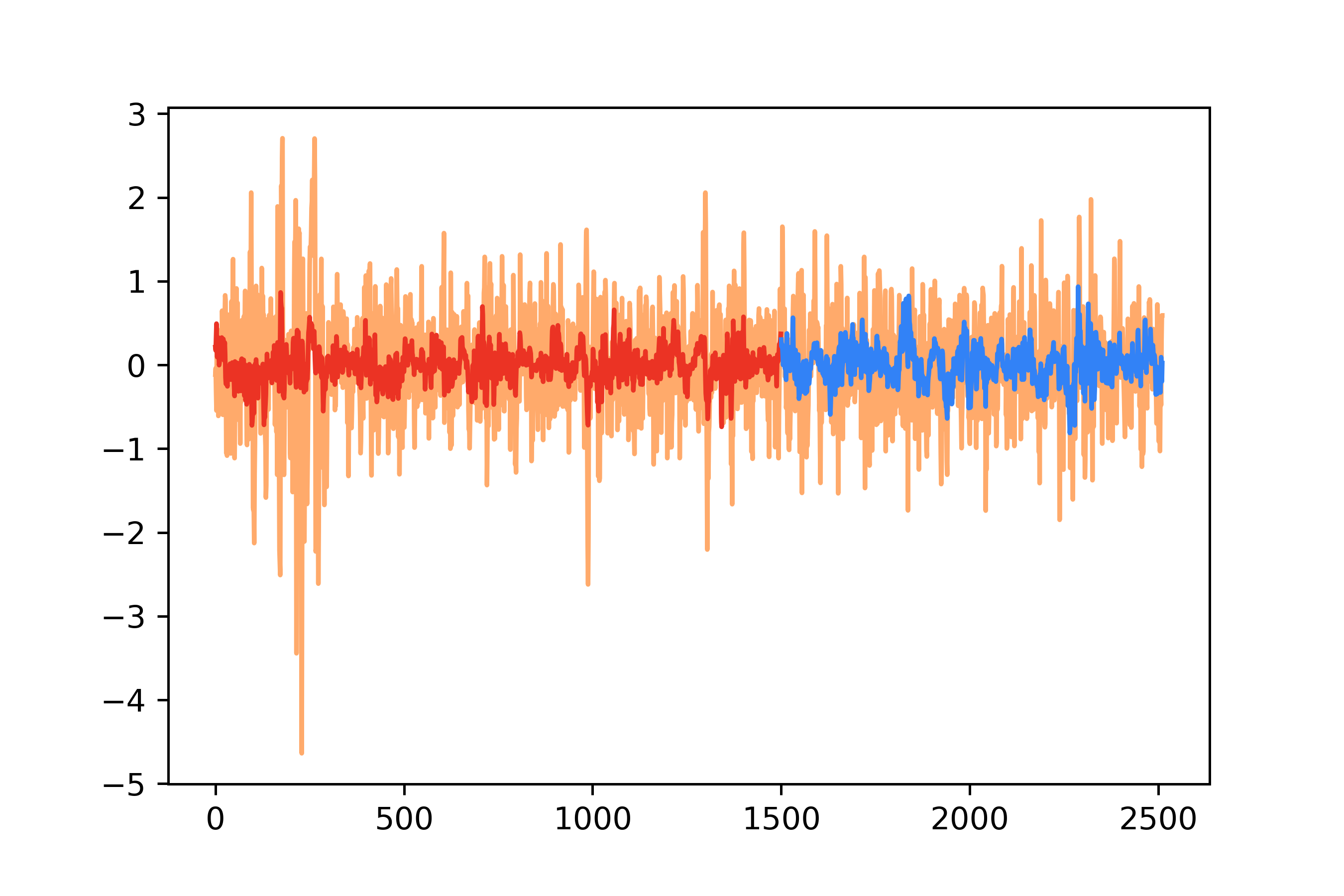}}{\scriptsize $\dot{\omega_x}$} &
            \stackunder{\includegraphics[width=0.16\textwidth]{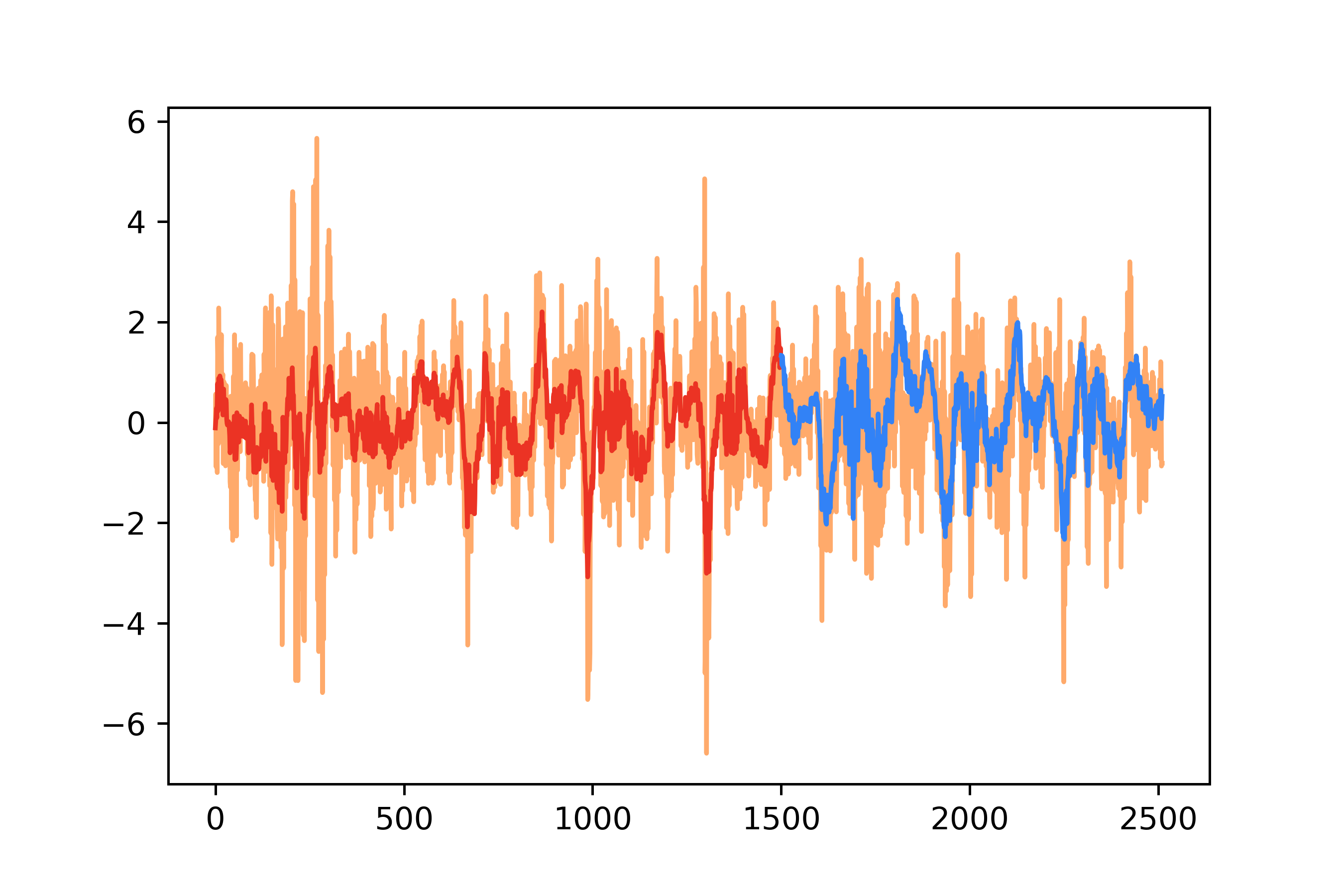}}{\scriptsize $\dot{\omega_y}$} &
            \stackunder{\includegraphics[width=0.16\textwidth]{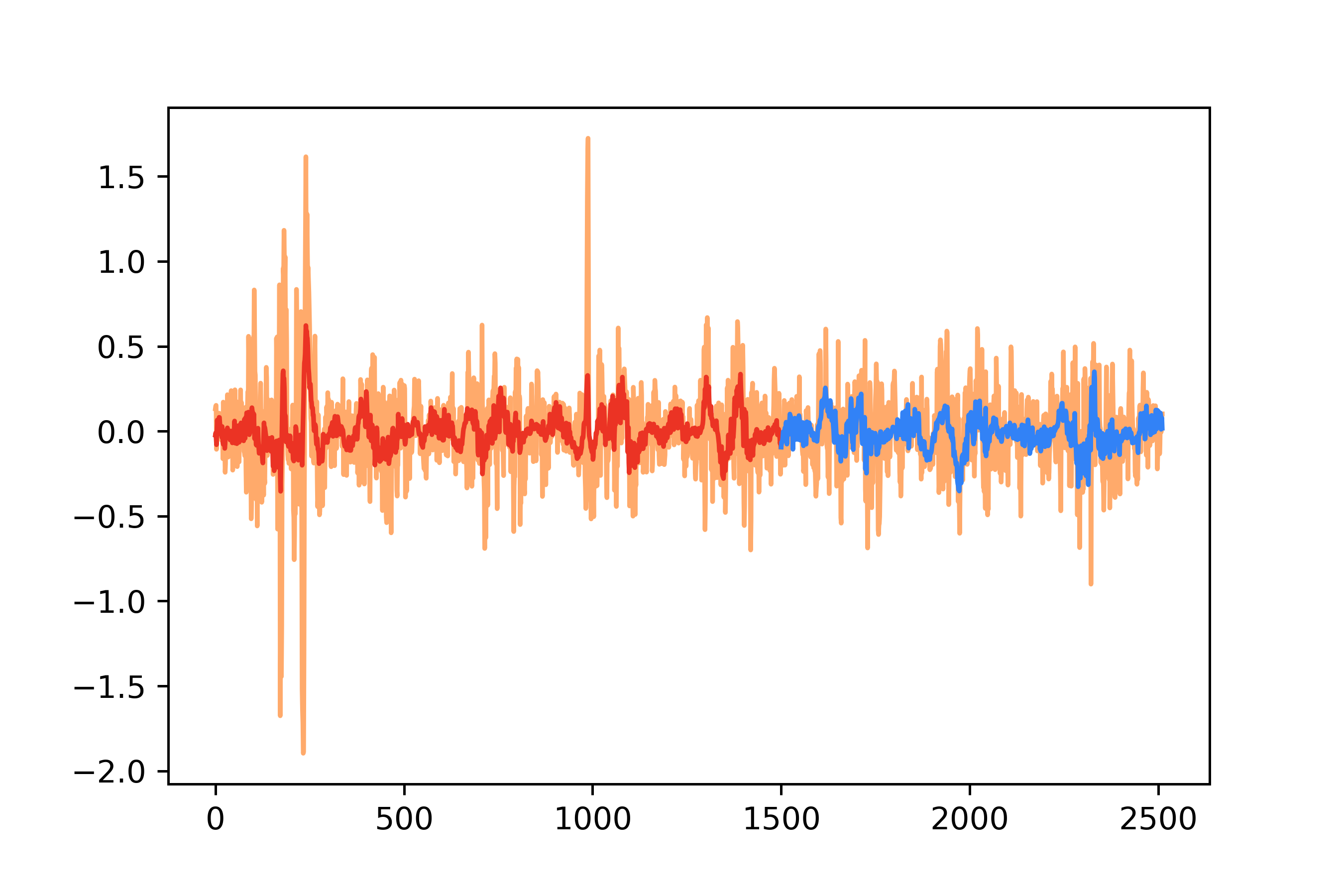}}{\scriptsize $\dot{\omega_z}$}
        \end{tabular} \\ \hline
    \end{tabular}
    \caption{\small Qualitative results of the full dynamics modeled by SINDy 
    and LeARN across different wind conditions. The horizontal axis runs 
    across 2511 data points collected over a time period of 50.2s sampled at 0.02s and the vertical axis represents the value.}
    \label{fig:full_qual}
\end{figure*}

\subsection{Comparative Performance Analysis}

For the analysis of adaptation and generalization performance, we evaluate the mean-squared error (MSE) between the ground truth and the dynamical predictions. Throughout, \textit{adaptation performance} refers to the MSE 
evaluated on the support data used for online gradient updates, 
and \textit{generalization performance} refers to the MSE on 
held-out query data from the same task on which no gradient 
updates are performed. 
Despite the limitations discussed above, we observe that LeARN delivers a much more competitive generalization performance as the input feature space begins to increase in dimension even with a light-weight DNN. It can be observed that the average difference in the generalization performance across the different adaptation wind tasks reduces from 0.043737 in Tables~\ref{tab:translational_comparison} to 0.002853 in Table~\eqref{tab:attitude_comparison}, and further down to 0.001811 in Table~\ref{tab:full_performance_comparison}, as the dimensionality of the concatenated input feature $X$ increases. We use the PySINDy~\cite{pysindy} package to model the SINDy dynamics representations and the Higher~\cite{grefenstette2019generalized} package to implement the bi-level MAML training loop.

\begin{table}[h]
\centering
\caption{Translational Dynamics Estimation Error: SINDy vs. LeARN}
\label{tab:translational_comparison}
\small
\begin{tabular}{l cccc}
\toprule
\multirow{2}{*}{\textbf{Condition}} & \multicolumn{2}{c}{\textbf{SINDy}} & \multicolumn{2}{c}{\textbf{LeARN (Ours)}} \\ 
\cmidrule(r){2-3} \cmidrule(l){4-5}
& \textit{Adapt.} & \textit{Gen.} & \textit{Adapt.} & \textit{Gen.} \\ 
\midrule
35wind   & 0.156 & \textbf{0.194} & \textbf{0.155} & 0.208 \\ 
70psin20 & \textbf{0.138} & \textbf{0.105} & 0.157 & 0.149 \\ 
70wind   & \textbf{0.193} & \textbf{0.172} & 0.210 & 0.219 \\ 
100wind  & \textbf{0.227} & \textbf{0.232} & 0.268 & 0.302 \\ 
\bottomrule
\end{tabular}
\end{table}

We present a quantitative comparison of the partial translational dynamics learnt by utilizing the formulation in Eq~\ref{eq:partial_dynamics}, by parameterizing $f(\cdot)$ through the LeARN framework in Table~\ref{tab:translational_comparison}.

\begin{table}[h]
\centering
\caption{Attitude Dynamics Estimation Error: SINDy vs. LeARN}
\label{tab:attitude_comparison}
\small
\begin{tabular}{l cccc}
\toprule
\multirow{2}{*}{\textbf{Condition}} & \multicolumn{2}{c}{\textbf{SINDy}} & \multicolumn{2}{c}{\textbf{LeARN (Ours)}} \\ 
\cmidrule(r){2-3} \cmidrule(l){4-5}
& \textit{Adapt.} & \textit{Gen.} & \textit{Adapt.} & \textit{Gen.} \\ 
\midrule
35wind   & 0.425 & 0.467 & \textbf{0.415} & \textbf{0.461} \\ 
70psin20 & \textbf{0.578} & \textbf{0.458} & 0.629 & 0.493 \\ 
70wind   & 0.563 & \textbf{0.827} & \textbf{0.529} & 0.830 \\ 
100wind  & 0.993 & 1.115 & \textbf{0.993} & \textbf{1.094} \\ 
\bottomrule

\end{tabular}
\end{table}

The partial attitude dynamics is learnt by utilizing the formulation in Eq~\ref{eq:attitude_partial_dynamics}. A quantitative comparison is presented in Table~\ref{tab:attitude_comparison}.
\begin{table}[h]
\centering
\caption{Full Dynamics Estimation Error: SINDy vs. LeARN}
\label{tab:full_performance_comparison}
\small
\begin{tabular}{l cccc}
\toprule
\multirow{2}{*}{\textbf{Condition}} & \multicolumn{2}{c}{\textbf{SINDy}} & \multicolumn{2}{c}{\textbf{LeARN (Ours)}} \\ 
\cmidrule(r){2-3} \cmidrule(l){4-5}
& \textit{Adapt.} & \textit{Gen.} & \textit{Adapt.} & \textit{Gen.} \\ 
\midrule
35wind   & 0.307 & 0.339 & \textbf{0.264} & \textbf{0.318} \\ 
70psin20 & 0.339 & \textbf{0.273} & \textbf{0.336} & 0.302 \\ 
70wind   & \textbf{0.375} & \textbf{0.482} & 0.379 & 0.486 \\ 
100wind  & \textbf{0.632} & 0.726 & 0.645 & \textbf{0.722} \\ 
\bottomrule
\end{tabular}
\end{table}

Table~\ref{tab:full_performance_comparison} presents the quantitative comparison for full quadrotor dynamics. The full quadrotor dynamics is learnt by utilizing the formulation in Eq~\ref{eq:full_dynamics}. Here, the full-dynamics mapping $h(\cdot)$ is factored into the following parameterized learnable components:
\begin{equation}
h(X) \approx \Theta(X;\psi) \mathcal{E}(X;\phi)^{T},
\label{eq:factorization}
\end{equation}
where $X$ is obtained by concatenating \begin{equation}
\begin{bmatrix}
v \\ 
\omega 
\end{bmatrix} \text{and } u = \begin{bmatrix}
n_1^2 \\ 
n_2^2 \\
n_3^2 \\
n_4^2
\end{bmatrix}.
\end{equation}
In addition, we present a qualitative side-by-side comparison of LeARN against SINDy in different evaluation wind conditions in Fig.~\ref{fig:full_qual}.


\section{Conclusion}
\label{sec:conclusion}

In this work, we introduced LeARN, a novel algorithm for meta-learning the basis functions for nonlinear system identification. Our proposed approach demonstrates significant adaptability and generalization capabilities in modeling dynamical systems, as evidenced by its performance on unseen wind conditions in the Neural Fly dataset. Unlike the SINDy algorithm, our approach learns the library of basis functions directly from the data. The proposed algorithm yields a parameterized basis function library optimized for adaptability to new, unseen environments. This is intended to improve robustness without requiring system-specific redesigns when encountering new dynamic regimes. Notably, we demonstrated that LeARN significantly narrows the performance gap with SINDy, particularly as the dimensionality of the input feature space increases, underscoring its suitability for capturing higher dimensional, complex nonlinear dynamical interactions. In the future, we aim to explore higher dimensional input feature spaces and techniques for learning basis functions for modeling intrinsic residual dynamics. 

Looking ahead, our study sets the groundwork for addressing the broader goal of enabling robots to autonomously model and adapt to dynamic unstructured environments. Furthermore, our approach paves the way for developing robust, versatile, and adaptive robotic systems capable of operating safely and effectively in the real-world where such approaches can be utilized for health monitoring of these systems via data-driven system identification.

\end{document}